\documentclass[journal]{IEEEtran}

\usepackage{cite}

\ifCLASSINFOpdf
\usepackage{amssymb}
\else
\fi
\usepackage{amsfonts}
\usepackage{amsmath}
\usepackage{url}
\usepackage{booktabs}
\usepackage{graphicx}
\usepackage{tabularray}
\usepackage{hyperref}
\usepackage{multirow}
\usepackage{orcidlink}
\usepackage{graphicx}
\usepackage{academicons}

\usepackage{color}

\usepackage{algorithm}
\usepackage{algpseudocode}

\title{BihoT: A Large-Scale Dataset and Benchmark for Hyperspectral Camouflaged Object Tracking}

\author{Hanzheng~Wang\orcidlink{0000-0001-7276-3216},
        Wei~Li\orcidlink{0000-0002-8599-3632}, \textit{Senior Member, IEEE,}
        Xiang-Gen Xia\orcidlink{0000-0002-5599-7683}, \textit{Fellow, IEEE,}
        and Qian~Du\orcidlink{0000-0001-8354-7500}, \textit{Fellow, IEEE.}

\thanks{This work is supported by NSFC Projects of International Cooperation and Exchanges under Grant W2411055 and the National Natural Science Foundation of China under Grant 62201043. (Corresponding author: Wei Li.)

Hanzheng Wang and Wei Li are with the School of Information and Electronics, Beijing Institute of Technology, the Beijing Key Laboratory of Fractional Signals and Systems, and also with National Key Laboratory of Science and Technology on Space-Born Intelligent Information Processing, Beijing 100081, China (e-mail: hzwangc@bit.edu.cn; liwei089@ieee.org).

Xiang-Gen Xia is with the Department of Electrical and Computer Engineering, University of Delaware, Newark, DE 19716, USA (e-mail: xxia@ee.udel.edu).

Qian Du is with the Department of Electrical and Computer Engineering, Mississippi State University, Starkville, MS 39762, USA (e-mail: du@ece.msstate.edu).}}

\begin{document}


\maketitle
%

\begin{abstract}
Hyperspectral object tracking (HOT) has many important applications, particularly in scenes where objects are camouflaged. The existing trackers can effectively retrieve objects via band regrouping because of the bias in the existing HOT datasets, where most objects tend to have distinguishing visual appearances rather than spectral characteristics. This bias allows a tracker to directly use the visual features obtained from the false-color images generated by hyperspectral images without extracting spectral features. To tackle this bias, the tracker should focus on the spectral information when object appearance is unreliable. Thus, we provide a new task called hyperspectral camouflaged object tracking (HCOT) and meticulously construct a large-scale HCOT dataset, BihoT, consisting of 41,912 hyperspectral images covering 49 video sequences. The dataset covers various artificial camouflage scenes where objects have similar appearances, diverse spectrums, and frequent occlusion, making it a challenging dataset for HCOT. Besides, a simple but effective baseline model, named spectral prompt-based distractor-aware network (SPDAN), is proposed, comprising a spectral embedding network (SEN), a spectral prompt-based backbone network (SPBN), and a distractor-aware module (DAM). Specifically, the SEN extracts spectral-spatial features via 3-D and 2-D convolutions to form a refined prompt representation. Then, the SPBN fine-tunes powerful RGB trackers with spectral prompts and alleviates the insufficiency of training samples. Moreover, the DAM utilizes a novel statistic to capture the distractor caused by occlusion from objects and background, and corrects the deterioration of the tracking performance via a novel motion predictor. Extensive experiments demonstrate that our proposed SPDAN achieves state-of-the-art performance on the proposed BihoT and other HOT datasets.
\end{abstract}



\section{Introduction}

\begin{figure} [!htb] 
\includegraphics[width=7.0cm,height=7.5cm]{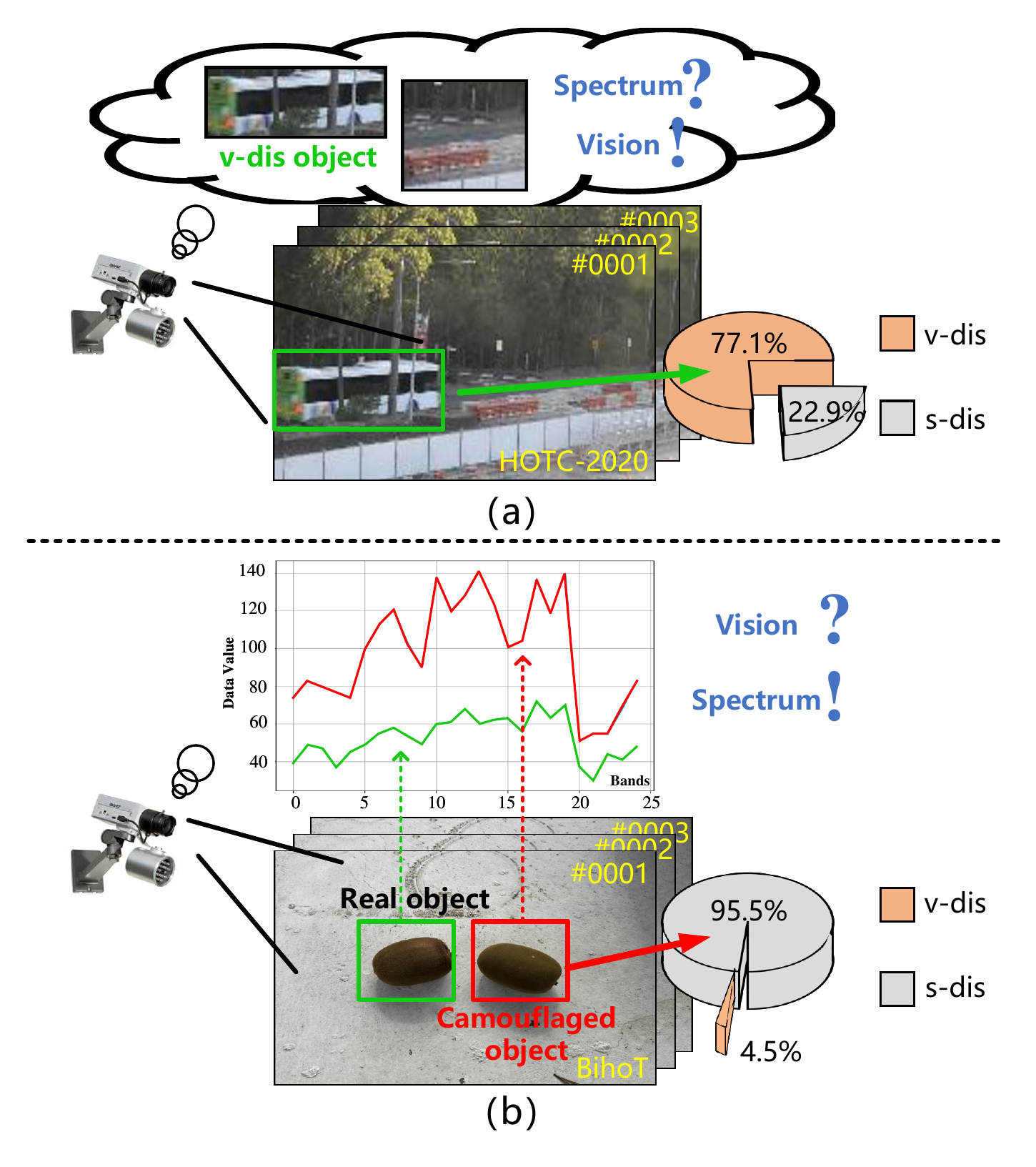}
\centering
\caption{Differences between the BihoT dataset and the HOTC-2020 dataset. The object in the green box is a real kiwi, while the object in the red box is a fake kiwi, considered a camouflaged object. Data value refers to the value of a pixel, representing the intensity of the spectral reflectance curve.}
\label{fig:1}
\end{figure}


\IEEEPARstart{V}{isual} object tracking \cite{jiao2021deep, nai2021robust} is a crucial task in many applications, such as autonomous driving, video analysis, etc. \cite{thayalan2023multifocus, tian2020simultaneous, bi2022multiscale}. It is to find the same objects across consecutive frames based on their initial positions in the first frame. However, it faces challenges with interference from similar appearances and background clutter. Hyperspectral images (HSIs), which provide both texture and spectral information, can effectively track objects despite visual interference, which is widely used in remote sensing fields \cite{hong2024spectralgpt, wang2023spectral}. Thus, hyperspectral object tracking (HOT) has garnered increased attention from researchers.

With the construction of HOT datasets such as HOT-2020 \cite{xiong2020material}, IMEC-25 \cite{chen2021object}, and WHU-Hi-H$^3$ \cite{liu2021unsupervised}, many deep learning-based methods have been proposed \cite{wu2024domain, chen2024sense, chen2023spirit, liu2022siamhyper, sun2023siamohot}. While these methods have demonstrated substantial progress, many discriminative spectral features remain under-explored. As illustrated in Fig. \ref{fig:1}(a), we identify two key characteristics in the existing HOT datasets: 1) the visually distinguishable (v-dis) factor, which indicates that objects can be accurately tracked using visual information, and 2) the spectral distinguishable (s-dis) factor, which reflects the ability to track objects based on spectral information. Taking the HOTC-2020 dataset as an example, video sequences with the v-dis factor account for 77.1\% of the entire dataset, while those with the s-dis factor only account for 22.9\%. Considering that most existing HOT methods rely on pre-trained RGB trackers for processing false-color images generated from HS images \cite{chen2024sense, chen2023spirit, li2023learning}, which are good at handling v-dis factor, even when a tracker struggles to extract reliable spectral features, it can still identify objects by leveraging visual cues (v-dis factor). This bias of datasets makes most existing algorithms overlook the advantages of hyperspectral (HS) information.

Despite the existing HOT tasks exploring the degradation of object visual information due to factors like illumination changes and motion blur, objects typically remain in a v-dis state. This condition prevents trackers from fully leveraging the spectral characteristics of objects, thereby limiting their ability to handle complex tracking environments. To address this challenge, we introduce a novel task, Hyperspectral Camouflaged Object Tracking (HCOT), which focuses on tracking objects that share the same visual information but differ in spectral characteristics. This technology enhances the focus on tracked objects while mitigating interference from artificial decoys, thereby improving the accuracy of criminal tracking and the reliability of security systems. Considering that no existing dataset suits this task, we construct the first large-scale HCOT dataset, BihoT. This dataset offers two significant advantages over existing datasets. First, BihoT collects numerous visually similar objects—camouflaged objects. As illustrated in Fig. \ref{fig:1}(b), the plastic fake fruit is a camouflaged object relative to the real fruit. In BihoT, the proportion of the s-dis factor reaches 95.5\%, while the proportion of the v-dis factor is only 4.5\%. For camouflaged objects, the v-dis factor is typically unreliable in tracking, underscoring the importance of the s-dis factor. Second, BihoT comprises 49 hyperspectral video sequences (HSVs) containing 41,912 HSIs in total, making it the largest HCOT dataset with 25 bands. Each HS image encompasses 25 bands and a resolution of 409 $\times$ 217, meeting the data requirements for deep learning.

Furthermore, to promote the development of HCOT, we propose a spectral prompt-based distractor-aware network (SPDAN), a simple yet effective baseline designed to address the challenges of spectral feature extraction and frequent occlusion. Existing SOTA HS trackers merely adjust the number of input channels in existing models, such as ResNet, without effectively extracting fine-grained information from the spectral dimension \cite{liu2022siamhyper, wu2024domain}. Thus, a spectral embedding network (SEN) is designed to transform the spatial and spectral information into tokenized features via a 3-D convolutional layer. The 3-D convolutions simultaneously establish spectral and spatial correlations across multiple bands, allowing it to capture pixel variations between different bands and extract rich spectral information \cite{roy2023multimodal, sun2022spectral, roy2023spectral}. Then, motivated by prompt learning \cite{zhu2023visual}, a spectral prompt-based backbone network (SPBN) is proposed to transfer knowledge from the RGB domain to the HS domain, enabling robust spectral feature extraction. To tackle frequent occlusions in practical applications, we introduce a distractor-aware module (DAM) to capture frames with disturbances such as occlusion accurately. Different from \cite{li2024object}, DAM considers both APCE and the offset of the prediction boxes across frames, which can make more accurate judgments on distractors. Moreover, it incorporates motion state models to predict object states during occlusion, effectively mitigating the decline in tracking performance caused by such distractors. The main contributions of this paper are as follows:

\begin{enumerate}
\item The first large-scale HCOT dataset, BihoT, is constructed, which contains 41,912 HS camouflaged images paired with the bounding box annotations. This dataset introduces more challenging scenarios and provides valuable insights for advancing HOT research.

\item A novel spectral embedding network is designed to transfer the spatial and spectral information to the tokenized features. SEN effectively aggregates spectral and visual features through prompt-based learning, enhancing the representation of camouflaged objects.

\item A distractor-aware module is proposed to perceive the distractor by considering both the average peak-to-correlation energy and the offset of the predicted bounding boxes. DAM rectifies tracking results using a motion predictor, improving the robustness against distractors.

\end{enumerate}

The remaining of this paper is organized as follows. Section II reviews some related works. Section III introduces the proposed BihoT dataset. Section IV formulates the proposed methods. Section V provides the ablation and comparative experimental results. Finally, conclusions are drawn in Section VI.

\begin{table*}[]
\caption{comparison of existing HOT benchmark datasets. The DATASET INFORMATION is listed, including YEAR, band number, training number (training num.), testing number (testing num.), Labeled sequence number (Labeled Sequence Num.), Average number of images per sequence (Avg. \#images/sequences), and annotation number (annotations).}
\label{tab:1}
\centering
\renewcommand{\arraystretch}{1.2}
\scalebox{1.0}{
\begin{tabular}{cccccccc}
\hline
Datasets        & Year & Band     & Training Num. & Testing Num. & Labeled Sequence Num. & Avg. \#images/sequences  &  Annotations \\ \hline
HOTC-2020\cite{xiong2020material}        & 2020 & 16       & 12,839        & 16,574       & 75                    & 392.17                  & 29,413     \\
IMEC-25\cite{chen2021object}         & 2021 & 25       & 14,062        & 14,357       & 135                   & 210.51                  & 28,419     \\
WHU-Hi-H$^3$\cite{liu2021unsupervised}       & 2022 & 25       & 22,753        & 6969         & 9                     & 774.33                  & 6969    \\
HOTC-2023-RedNIR\cite{HOTC-2023} & 2023 & 15       & 5737          & 3500         & 14                    & 659.78                  & 9237    \\
HOTC-2023-VIS\cite{HOTC-2023}    & 2023 & 16       & 19,228        & 20,074       & 55                    & 714.58                  & 39,302   \\
HOTC-2023-NIR\cite{HOTC-2023}    & 2023 & 25       & 8161          & 4856         & 40                    & 325.42                  & 13,017    \\ \hline
BihoT           & 2024 & 25       & 22,992        & 18,920       & 49                    & 855.35                  & 41,912    \\ \hline
\end{tabular}%
}
\end{table*}

\section{Related work}

\subsection{HS Object Tracker}
HOT aims to discover the same objects in subsequent frames based on the initial object state and spectral information. Early works mostly used the traditional methods for predicting the object position, such as histogram feature-based methods \cite{xiong2020material}, correlation filters-based methods \cite{tang2022target, uzkent2018tracking}, and spectral reflectance-based methods \cite{van2010tracking}, which has a limited ability to deal with the complicated distractors.

Many deep learning-based methods have been recently proposed to extract robust deep features and handle complex scenes such as occlusion. It can be mainly categorized into band selection-based methods and feature fusion-based methods. The former methods leverage the RGB tracker to predict the motion of objects via the false-color images generated from HSIs. For example, Li \textit{et al.} \cite{li2023learning} proposed an ensemble learning-based method based on the false-color images generated by a band-selection method. The network produced the band weights via an auto-encoder-like structure, which is used to regroup multiple bands to several three-channel false-color images. Li \textit{et al.} \cite{li2023siambag} designed an attention-based band grouping method to generate the importance weights of different bands. Chen \textit{et al.} \cite{chen2024sense} proposed a spectral-spatial self-expression module to transform the HSI into false-color images by filling the modality gap. Chen \textit{et al.} \cite{chen2023spirit} presented a spectral awareness module to explore the band weight distributions by the multi-level band interactions. Methods based on band selection can leverage the processing capabilities of RGB trackers to predict object positions from three-channel images. However, this is due to the inherent bias in existing datasets. The v-dis factor of HS objects collected in natural scenes tends to dominate, while the s-dis factor is often overlooked. As a result, the tracking algorithm is not effectively guided to explore the richer spectral information available in HS data, which fails to meet practical application requirements.

The latter methods mainly adopt the Siamese tracker framework, comprising the spectral and RGB tracking branches to mine the cross-modality semantic cues. For example, Liu \textit{et al.} \cite{liu2022siamhyper} presented a spectral and visual feature fusion framework SiamHYPER to provide a rich semantic feature representation and obtained accurate tracking results from the cross-modality complementary decision. Wu \textit{et al.} \cite{wu2024domain} proposed a Siamese Transformer tracker to learn the heterogeneous features from different HSIs by the domain label reverse learning. Jiang \textit{et al.} \cite{jiang2024channel} proposed a unified network, termed as SiamCAT, to process HSIs that have varied numbers of bands. Besides, with a guided learning attention module, the spectral features can be highlighted by the spatial attention obtained from the RGB branch. Li \textit{et al.} \cite{li2024material} designed a material-guided multiview fusion network MMF-Net, which combines the RGB, spectral, and material information to form a unified rich representation of an object. However, most existing methods simply adjust the number of input channels in the network to process HS data, resulting in limited spectral feature extraction. To address the above two issues, we construct a camouflaged object tracking dataset and propose a baseline model called SPDAN. SPDAN first utilizes a novel spectral embedding network to extract HS features. Then, a distractor-aware module is proposed to consider the position response from both the deep model and the Kalman filter, achieving robust tracking performance.

\subsection{HOT Datasets}

The existing HOT datasets are listed in Table \ref{tab:1}, most containing 16 or 25 bands. The detailed information is introduced as follows.

1) \textit{HOTC-2020} \cite{xiong2020material} is the first mainstream 16-band HOT dataset, consisting of 75 HSV sequences, of which 40 are used for training and 35 are used for testing, totaling 29,413 HSIs. The main shooting objects include cars and pedestrians.

2) \textit{IMEC-25} \cite{chen2021object} consists of 80 HSV sequences for training and 55 HSV sequences for testing. The main shooting objects include cars, pedestrians, and planes, which are collected by a 25-band IMEC camera.

3) \textit{WHU-Hi-}H$^3$ \cite{liu2021unsupervised} contains 22,753 25-band frames of annotated HSVs covering nine scenes, where 6969 frames are for training, and the remaining images are for testing. The main shooting objects include cars and pedestrians.

4) \textit{HOTC-2023} is available at the HOT competition website and consists of three sub-datasets: 15-band, 16-band, and 25-band sets, which are collected from wavelength ranges of 600-850, 460-600, and 665-960 nm, respectively. The main shooting objects include cars and pedestrians.

It can be seen that the above datasets neglect the camouflage characteristics of objects and only collect images in natural scenes, with most objects being vehicles and pedestrians, lacking the challenge of visual similarity. Although the WHU-Hi-H$^3$ dataset considers the similarity between objects relative to the background, it does not highlight the visual confusion caused by the similarity between objects.

\subsection{Visual Object Tracker}

In recent years, the Transformer has been applied in many computer vision tasks such as detection \cite{fan2019shifting}, segmentation \cite{strudel2021segmenter, zhou2022survey}, and multi-object tracking \cite{gu2022rpformer,gu2024rtsformer,gu2023eantrack} due to its outstanding ability to model long-term dependencies. Many Transformer-based tracking methods have been proposed. Xie \textit{et al.} \cite{xie2022correlation} proposed a novel object-dependent feature network to embed the inter-image correlation. The non-object features can be suppressed in the self- and cross-attention schemes. Song \textit{et al.} \cite{song2022transformer} designed a multi-scale cyclic shifting window attention to capture the window-level features. Mayer \textit{et al.} \cite{mayer2022transforming} proposed a Transformer-based model prediction module, which captures global correlations without inductive bias. Cui \textit{et al.} \cite{cui2024mixformer} presented a compact tracking framework MixFormer, which extracts and fuses features simultaneously through a mixed attention module without the need for additional fusion modules or box post-processing. Zhu \textit{et al.} \cite{zhu2023visual} developed a visual prompt tracking framework to achieve task-oriented multi-modality tracking. Zhou \textit{et al.} \cite{zhou2022survey} presented a detailed overview of the representative literature on both tracking methods and datasets. With the guidance of learned prompts, existing RGB-domain foundation models can effectively handle the downstream multi-modality tracking tasks. For HS trackers, there is currently limited research based on the Transformers.

\subsection{RGB/RGBT Tracking Datasets and Evaluation Metrics}

The development of object tracking tasks has benefited from the promotion of a large number of publicly available datasets, particularly in the domains of RGB and RGB-T object tracking. RGB object tracking datasets are primarily used to evaluate tracking algorithms in visible conditions. Among them, the OTB series (OTB-2013\cite{wu2013online}, OTB-2015\cite{wu2015tracking}) is one of the earliest benchmark datasets, encompassing various challenging factors such as object deformation, lighting variations, and occlusions. It employs success plots and precision plots as evaluation metrics. The VOT Challenge introduces new datasets annually, such as VOT2016, VOT2018, and VOT2019, which adopt strict evaluation criteria, including overlap accuracy and robustness. Additionally, they incorporate a re-initialization mechanism, making them critical benchmarks for evaluating tracking algorithms in competitive settings. In recent years, large-scale, long-term tracking datasets such as LaSOT\cite{fan2019lasot} and TrackingNet\cite{muller2018trackingnet} have further advanced the field. LaSOT consists of 1,400 long-duration video sequences, each averaging over 2,500 frames, making it a valuable resource for studying long-term object tracking. TrackingNet, on the other hand, is sourced from YouTube and contains over 30,000 video sequences, spanning a diverse range of real-world scenarios.

RGB-T object tracking datasets offer greater environmental adaptability by integrating RGB and thermal information. GTOT \cite{li2016learning} is one of the earliest RGB-T tracking datasets, comprising 50 video sequences. The RGBT234 \cite{li2019rgb} dataset significantly expands this, containing 234 video sequences with a total frame count exceeding 234,000. It encompasses 12 challenging factors, including modality variations, camera motion, and object deformation, providing a more comprehensive benchmark for evaluating multi-modality object tracking algorithms. However, existing RGB-T datasets still face limitations in terms of diversity and scale. To address these shortcomings, the LasHeR \cite{li2021lasher} dataset has been introduced as the largest RGB-T object tracking benchmark. LasHeR consists of 1,224 video sequences with a total frame count exceeding 500,000, covering a wide range of scenarios such as indoor and outdoor environments, day and night conditions, and extended video durations, making it a crucial platform for studying long-term RGB-T tracking.

In recent years, many effective evaluation criteria have been proposed, including precision rate (PR), success rate (SR), accuracy, and Expected Average Overlap (EAO).

1) Precision rate is defined as the proportion of frames where the center location error (CLE) between the predicted bounding box and the ground truth remains below a specified threshold. The precision rate represents the percentage of such frames, while the precision plot illustrates how this rate varies as the threshold increases. Typically, a threshold of 20 pixels is used.

2) Success rate is determined by whether the overlap between the predicted bounding box and the ground truth exceeds a specified threshold. The success plot illustrates how the success rate varies as the threshold ranges from 0 to 1. To effectively compare different methods, the area under the curve (AUC) is used as a ranking metric.

3) Similar to the success rate, accuracy evaluates how well the predicted bounding box overlaps with the ground truth. It is determined by calculating the average overlap between the predicted and ground truth bounding boxes during successful tracking periods.

4) EAO systematically integrates per-frame accuracy and failure rates, providing a principled measure of a tracker's expected no-reset overlap on short-term sequences. While other evaluation metrics exist, Precision Rate (PR) and Success Rate (SR) remain the most commonly used in HOT \cite{liu2022siamhyper, li2023learning}, which is also what we use in our experiments.

\begin{figure*}
    \centering
    \includegraphics[width=16cm,height=7.0cm]{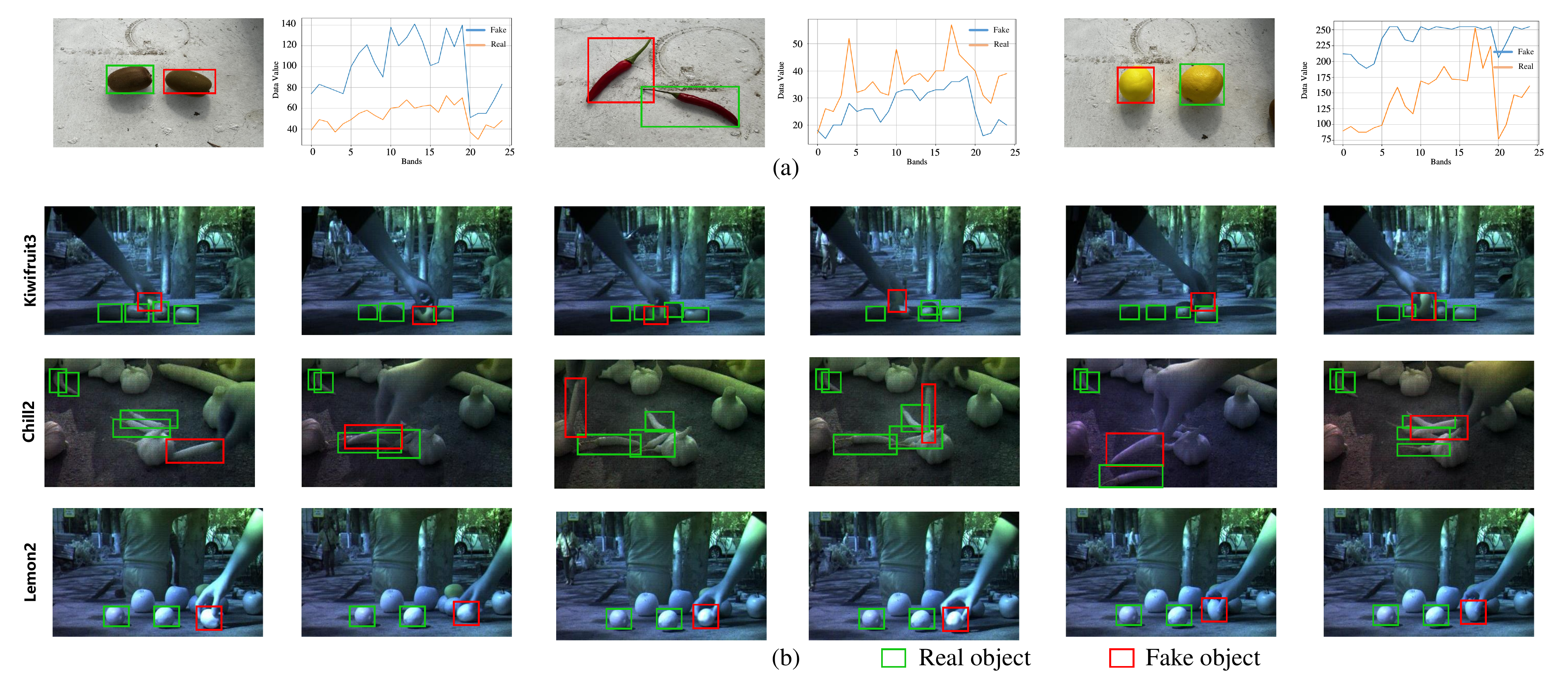}
    \caption{Illustration of the proposed BihoT dataset. (a) Examples of spectral distinguishable (s-dis) factors from the BihoT dataset. (b) Examples of false-color images of the \textit{Kiwifruit3}, \textit{Chill2}, and \textit{Lemon2} video sequences from the BihoT dataset.}
    \label{fig:3}
\end{figure*}

\section{BihoT Dataset}

\subsection{Image Collection}

We employ an IMEC XIMEA MQ022HG-IM-SM5X5-NIR camera and an EDMUND OPTICS 35mm C VIS-NIR lens to capture the high spatial and spectral resolution HS images across the wavelength range of 600-974 nm. This wavelength range includes a portion of the visible light spectrum. In this article, spectral information refers to the spectral information contained within the corresponding wavelength range of the dataset. The original image is a two-dimensional image with pixels of $2048\times1088$. After processing with the transformation algorithm, a 3D hyperspectral cube of $409\times217\times25$ can be obtained. To ensure the diversity of the scene, we collect data from different seasons and illuminations. The objects are real and artificial plastic fruits or vegetables in daily life to ensure the acquisition of camouflaged objects. which is shown in Fig. \ref{fig:3}. We define camouflage in two aspects. First, camouflage occurs when the object and background share identical colors, similar to camouflage object detection in the RGB domain. Second, camouflage exists between different objects, primarily characterized by objects with the same appearance but different materials. To support research in this area, we acquire various objects in daily life, including plastic and metal bowls, cups, plastic fruits, and more. This allows us to create diverse object combinations and conduct experiments under background clutter and camouflaged object interference conditions. In addition, BihoT also provides false-color images obtained by selecting combinations of the 1st, 9th, and 15th bands, which serve as RGB modality in our experiments.

\begin{table*}[]
\caption{Sequence names, attributes, and bounding box number of the testing set of the BihoT dataset.}
\label{tab:2}
\centering
\scalebox{1.0}{
\begin{tabular}{ccc|ccc}
\hline
Name    & Attribute                 & Box Num. & Name       & Attribute                 & Box Num. \\ \hline
Apple3  & IPR, IV, OCC, OPR, SV     & 770      & Kiwifruit1 & FM, IPR, IV, OCC, OPR, SC & 649      \\
Bottle1 & OCC, SV                   & 631      & Kiwifruit3 & FM, IPR, IV, OCC, OPR, SV & 885      \\
Bowel1  & IPR, OCC, OPR, SV         & 201      & Orange3    & BC, IV, OCC, SV           & 680      \\
Bowel2  & IPR, OCC, OPR, SV         & 248      & Paper      & BC, SV                    & 900      \\
Bowel3  & OCC, OPR, SV              & 102      & Tank1      & BC, SC                    & 820      \\
Chili1  & BC, IPR, LR, OCC, OPR, SC & 945      & Mango1     & BC, IPR, OCC, OPR, SC     & 870      \\
Chili2  & BC, LR, OCC, OPR, SV      & 1180     & Mango2     & BC, IPR, OCC, OPR, SV     & 830      \\
Cup1    & IV, SV                    & 1431     & Naloong2   & IPR, OCC, SV              & 1301     \\
Cup2    & IV, SV                    & 1666     & Lemon1     & FM, IPR, OCC, OPR, SC     & 750      \\
Gear2   & LR, OCC, OPR, SC, SV      & 1792     & Lemon2     & FM, IPR, OCC, OPR, SV     & 620      \\
Tank2   & BC, OPR, SV               & 1071     & Lemon3     & FM, IPR, IV, OCC, OPR, SV & 570      \\ \hline
\end{tabular}%
}
\end{table*}

During the image collection process, the location of the camera, exposure time, white balance, and other camera parameters are kept consistent. Besides, we found that some images do not contain the object, or the object moves out of the field of view during motion. Therefore, we carefully examine the entire dataset and filter out these noisy images.

\subsection{Data Annotation and Condition Challenges}
The BihoT dataset has 49 annotated sequences, of which 27 are for training, and 22 are for testing. Each image has an object marked by a bounding box. All bounding boxes were manually created using the Labelme toolkit. In addition, we set 9 attributes corresponding to different challenging factors in the testing set, including Background Clutter (BC), Fast Motion (FM), In-Plane Rotation (IPR), Illumination Variation (IV), Low Resolution (LR), Occlusion (OCC), Out-of-Plane Rotation (OPR), Spectral Consistency (SC), and Spectral Variation (SV). Table \ref{tab:2} lists the number of bounding boxes and challenging scenarios for each sequence. To measure the spectral feature extraction ability of the tracker, we proposed two challenging factors, Spectral Consistency (SC) and Spectral Variation (SV), compared with the existing datasets \cite{xiong2020material, liu2021unsupervised, chen2021object}. Specifically, in the BihoT dataset settings, there are multiple real objects and one camouflaged object in each sequence. SC means that when the real object moves, its spectral information remains unchanged relative to other real objects in the scene. When the camouflaged object moves, the spectral information relative to other real objects in the scene is different. Therefore, the scene where the camouflaged object moves is denoted as SV.

\subsection{Data Statistics}

Table \ref{tab:1} lists a statistics comparison between the proposed BihoT and the existing HOT datasets. We can see that BihoT is currently the largest 25-band HOT dataset, consisting of 41,912 images, with an average of 855.35 frames per video, making it a valuable dataset for developing HOT search.
 
\section{Proposed Method}

\begin{figure*}
    \centering
    \includegraphics[width=15cm,height=12.5cm]{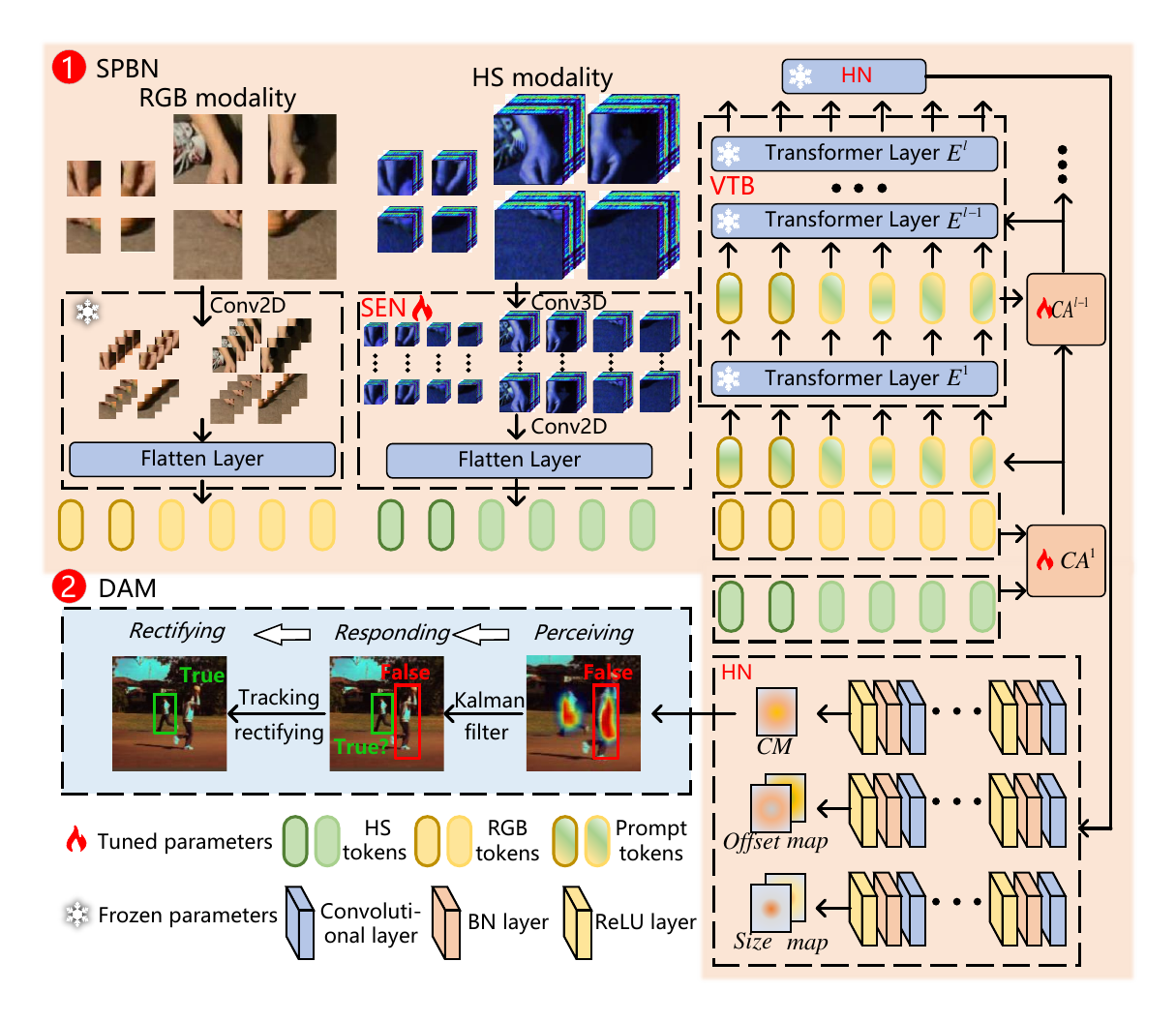}
    \caption{Illustration of the overall structure of our proposed SPDAN, including the spectral prompt-based backbone network (SPBN) and distractor-aware module (DAM). Specifically, SPBN contains three main modules, i.e., spectral embedding network (SEN), cross-modality adapter (CA), visual Transformer backbone (VTB), and head network (HN).}
    \label{fig:framework}
\end{figure*}

Fig. \ref{fig:framework} depicts the framework overview of the proposed SPDAN, which consists of two key components. \textit{1) Spectral Prompt-based Backbone Network:} This component designs a spectral embedding module that can help establish rich spectral embedding tokens and learn joint feature representations of HS and visual by embedding spectrum as prompts in the network. Therefore, it can inherit the powerful tracking ability of the visual foundation model while solving the problems of scarcity and poor portability of HS data. \textit{2) Distractor-aware Module:} This module solves disturbance interference by first establishing a perceptual disturbance statistic that can capture frames with occlusion or background interference in the video sequence. When the SPBN prediction results are unreliable, a motion model is used to update the object's position, which can effectively reduce the decrease in tracking performance caused by disturbances.

\subsection{Spectral Prompt-based Backbone Network}

Hyperspectral camouflaged object tracking task aims to search for the object in a continuous video sequence that is highly similar to the background and other objects. Camouflaged objects usually have the same appearance (color, shape, and scale) as the real ones, except for materials, so it is difficult to utilize the visual features to distinguish a camouflaged object. To overcome this issue, a spectral prompt-based backbone network (SPBN) is proposed, incorporating spectral information into the visual features to form an effective cross-modality feature representation.

The overall network structure is shown in Fig. \ref{fig:framework}. The SPBN consists of four components, including a spectral embedding network (SEN), a cross-modality adapter (CA), a visual Transformer backbone (VTB), and a head network (HN). The input of SPBN is cross-modality image pairs, i.e., an RGB image pair (template image $z_{rgb} \in {{\mathbb{R}}^{3 \times H_z \times W_z}}$ and search image $x_{rgb} \in {{\mathbb{R}}^{3 \times H_x \times W_x}}$) and an HS image pair (template image $z_{hs} \in {{\mathbb{R}}^{C \times H_z \times W_z}}$ and search image $x_{hs} \in {{\mathbb{R}}^{C \times H_x \times W_x}}$).  $H_z$ and $W_z$ denote the height and the width of a template image, and $H_x$ and $W_x$ denote the height and the width of a search image, respectively. $C$ denotes the number of bands of HS images. Different from existing methods based on Siamese networks \cite{li2023siambag, liu2022siamhyper}, our proposed SPBN can simultaneously extract features from cross-modality template images and search images and use the Transformer structure to implicitly model the relationship between template and search features, avoiding the extra relationship calculations in the Siamese network-based methods. 

1) Spectral Embedding Network (SEN). To mine the spectral information and perform the cross-modality feature representations, a spectral embedding network is designed. A 3-D convolution layer and a 2-D convolution layer are jointly used for embedding spectral information, which can extract spectral information while minimizing the increase in parameters. Taking the input HS image $x_{hs} \in {{\mathbb{R}}^{C \times H_x \times W_x}}$ as an example, the spectral feature extraction process for each spatial position $(\alpha, \beta, \gamma)$ of the $i$th feature  cube can be formulated by
\begin{equation}
u_i^{\alpha ,\beta ,\gamma } = f(\sum\limits_k {\sum\limits_{h = 0}^{H' - 1} {\sum\limits_{w = 0}^{W' - 1} {\sum\limits_{r = 0}^{R - 1} {\omega _{i,k}^{h,w,r}u_k^{\alpha  + h,\beta  + w,\gamma  + r}} } } }  + {b_i}),
\end{equation}
where $f( \cdot )$ denotes the activation function and $k$ is the feature cube from the previous layer. $H', W', R$ denote the height, width, and band number of the 3-D convolutional kernel, respectively. ${\omega _{i,k}^{h,w,r}}$ denotes the parameters of the position $(h, w, r)$, and $b_i$ is the bias. By doing so, the size $x_{hs}$ is first changed to $C \times D \times H \times W$. D denotes the depth dimension of 3-D convolution, which equals 16 in SPDAN. Due to its calculation with sliding windows in the spectral dimension, 3-D convolution can effectively extract spectral features. After obtaining the original spectral features, They are first split and flattened into sequences of patches ${x'_{hs}} \in {^{(C \times D) \times {H_x} \times {W_x}}}$, and a 2-D convolution layer encodes the spatial information further to form the final token. The process can be formulated by:
\begin{equation}
u_i^{\alpha ,\beta } = f(\sum\limits_k {\sum\limits_{h = 0}^{H'' - 1} {\sum\limits_{w = 0}^{W'' - 1} {w_{i,k}^{h,w}u_k^{\alpha  + h,\beta  + w}} } }  + {b_i}),
\end{equation}
where $H''$ and $W''$ denote the height and width of the 2-D convolutional kernel, respectively. ${w_{i,k}^{h,w}}$ denotes the parameters of the position $(h, w)$. In this way, the input $x'_{hs}$ can be embedded as a spectral token $x_{hs}^{token}\in {{\mathbb{R}}^{dim \times N_x}}$, where ${N_x} = {H_x}{W_x}/{P^2}$ is the number of tokens, $P$ is the downsampling factors, and $dim$ is the feature dimensions, which equals 768 in the proposed SPDAN.

\begin{figure}
    \centering
    \includegraphics[width=8.5cm,height=4.5cm]{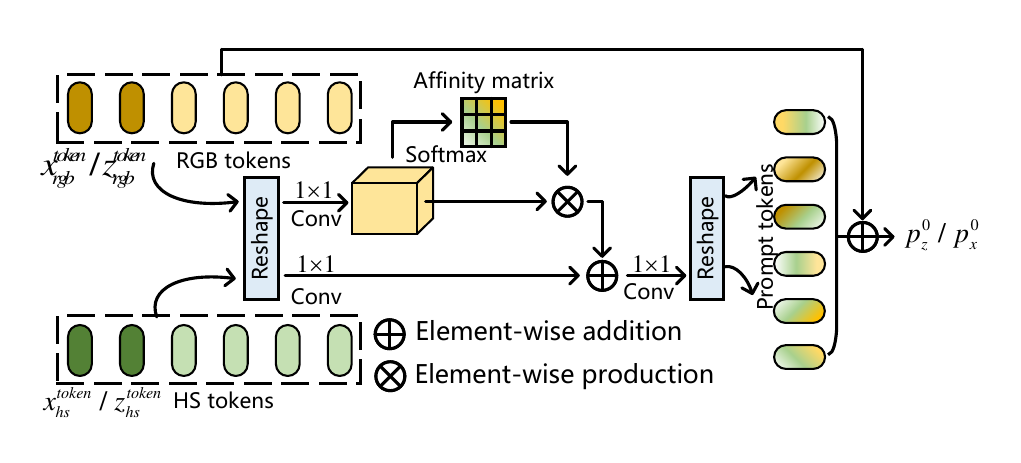}
    \caption{Illustration of the structure of CA.}
    \label{fig:CA}
\end{figure}

2) Cross-modality Adapter (CA). SPBN freezes the entire backbone model and only learns spectral prompts for a specific modality without training the visual Transformer backbone from scratch, enabling the model to converge quickly in the limited sample situations. A cross-modality adapter (CA) is employed to learn the hybrid prompt tokens from HS and visual modalities, which leverages the fovea attention to extract HS features. This integration allows the model to capture complementary information from both HS and visual modalities, enhancing the ability to distinguish fine-grained spectral details while maintaining spatial coherence. As shown in Fig. \ref{fig:CA}, the initial prompts can be obtained by $p_z^0 = CA(z_{rgb}^{token},z_{hs}^{token})$ and $p_x^0 = CA(z_{rgb}^{token},z_{hs}^{token})$. For the $l$th layer of the Transformer, the prompts learned from the CA can be formally represented as
\begin{equation}
\begin{array}{l}
p_z^l = {CA}^{l}(te_{outz}^{l - 1},p_z^{l - 1}), l=1,2,...,L,\\
p_x^l = {CA}^{l}(te_{outx}^{l - 1},p_x^{l - 1}), l=1,2,...,L,
\end{array}
\end{equation}
where $L>1$, $te_{out}^{l - 1}$ is the output feature of $(l-1)$th Transformer layer, and ${CA}^l$ denotes the $l$th cross-modality adapter. Specifically, a linear transformation is first used to reduce the dimension of cross-modality features to reduce feature redundancy. Then, to adaptively capture the complementary modality representations, a spatial attention mechanism is used to extract the complementary modality features. Finally, the features of the two modalities are mapped back to the input dimension to generate the cross-modality prompts. The overall process can be formulated by
\begin{equation}
{p^l} = pro{j_3}(SA(pro{j_1}(te_{out}^{l - 1})),pro{j_2}({p^{l - 1}})),
\end{equation}
\begin{equation}
{p^l} = concat(p_z^l,p_x^l),
\end{equation}
\begin{equation}
SA(m) = m \odot Softmax (Flatten(m)),
\end{equation}
where $pro{j_1}, pro{j_2}, pro{j_3}$ denote the different linear layers, respectively. $SA$ denotes the spatial attention mechanisms, and $\odot$ is the Hadamard product.

3) Visual Transformer Backbone (VTB). After the processing of the above two modules, the RGB and HS modalities can complement each other for prompt embedding, which can be input into a backbone model for end-to-end training. Here, we adopt the advanced visual tracking framework OSTrack \cite{ye2022joint}. Each embedded cross-modality prompt, including template and search images, is input into the Transformer encoder to extract features. It is worth noting that the parameters of this backbone model are frozen. We perform multi-stage modality embedding between each Transformer layer to extract semantic features from different stages. The forward propagation process can be formulated as
\begin{equation}
te_{out}^l = {E^l}({p^l}),l = 1,2,...,L,
\end{equation}
where $E^l$ denotes the $l$th Transformer encode layer, including multi-head self-attention sub-network, normalization layer, and feed-forward sub-network.

4) Head Network (HN). To obtain the precise position of an object, we design a head network to predict its classification score, position offset, and size, respectively. The overall structure is shown in Fig. \ref{fig:framework}. It consists of stacked convolutional layers followed by a batch normalization layer and an activation layer. The classification branch outputs a classification response map $CM \in {{\mathbb{R}}^{{\textstyle{{{H_x}} \over P}} \times {\textstyle{{{W_x}} \over P}}}}$, which is a 2-D matrix with a height of ${\textstyle{{{H_x}} \over P}}$ and a width of ${\textstyle{{{W_x}} \over P}}$, where ${CM}(i,j) \in [0,1]$ and ${CM}(i,j)$ denotes the response value at position $(i,j)$ of $CM$. Each point $(i, j)$ in $CM$ is a probability representing the foreground scores at the corresponding position of the search region. The regression branch outputs a normalized offset map $O \in {{\mathbb{R}}^{2 \times {\textstyle{{{H_x}} \over P}} \times {\textstyle{{{W_x}} \over P}}}}$ and a normalized bounding box size tensor $S \in {{\mathbb{R}}^{2 \times {\textstyle{{{H_x}} \over P}} \times {\textstyle{{{W_x}} \over P}}}}$, both of which are 3-D tensors with a depth of $2$, a height of ${\textstyle{{{H_x}} \over P}}$ and a width of ${\textstyle{{{W_x}} \over P}}$, where ${H_x}$ and ${W_x}$ denote the height and width of the input search image, respectively, and $P$ denotes a down-sampling ratio of the SPBN. Each point $(:, i, j)$ in $O$ and $S$ is a 2-D matrix indicating the offset and size of the box, respectively.

The position of the object is where the maximum value in $CM$ appears, and it can be expressed as $({i_p},{j_p}) = \arg {\max _{(i,j)}}{{CM}(i,j)}$. The final bounding box can be obtained by
\begin{equation}
\begin{array}{l}
(x,y,w,h) = ({x_p} + O(0,{i_p},{j_p}),{j_p} + O(1,{i_p},{j_p}),\\
S(0,{i_p},{j_p}),S(1,{i_p},{j_p})).
\end{array}
\end{equation}

To effectively train the SPDAN, we adopt the weighted focal loss \cite{law2018cornernet} for classification, and IoU loss \cite{rezatofighi2019generalized} and $L_1$ loss for regression
\begin{equation}
L = {L_{cls}} + {\lambda _1}{L_{iou}} + {\lambda _2}{L_1},
\end{equation}
where ${\lambda _1}$ and ${\lambda _2}$ are set to 2 and 5, respectively, according to \cite{yan2021learning}.

\begin{figure*} [!htb] 
\centering
\includegraphics[width=11cm,height=6cm]{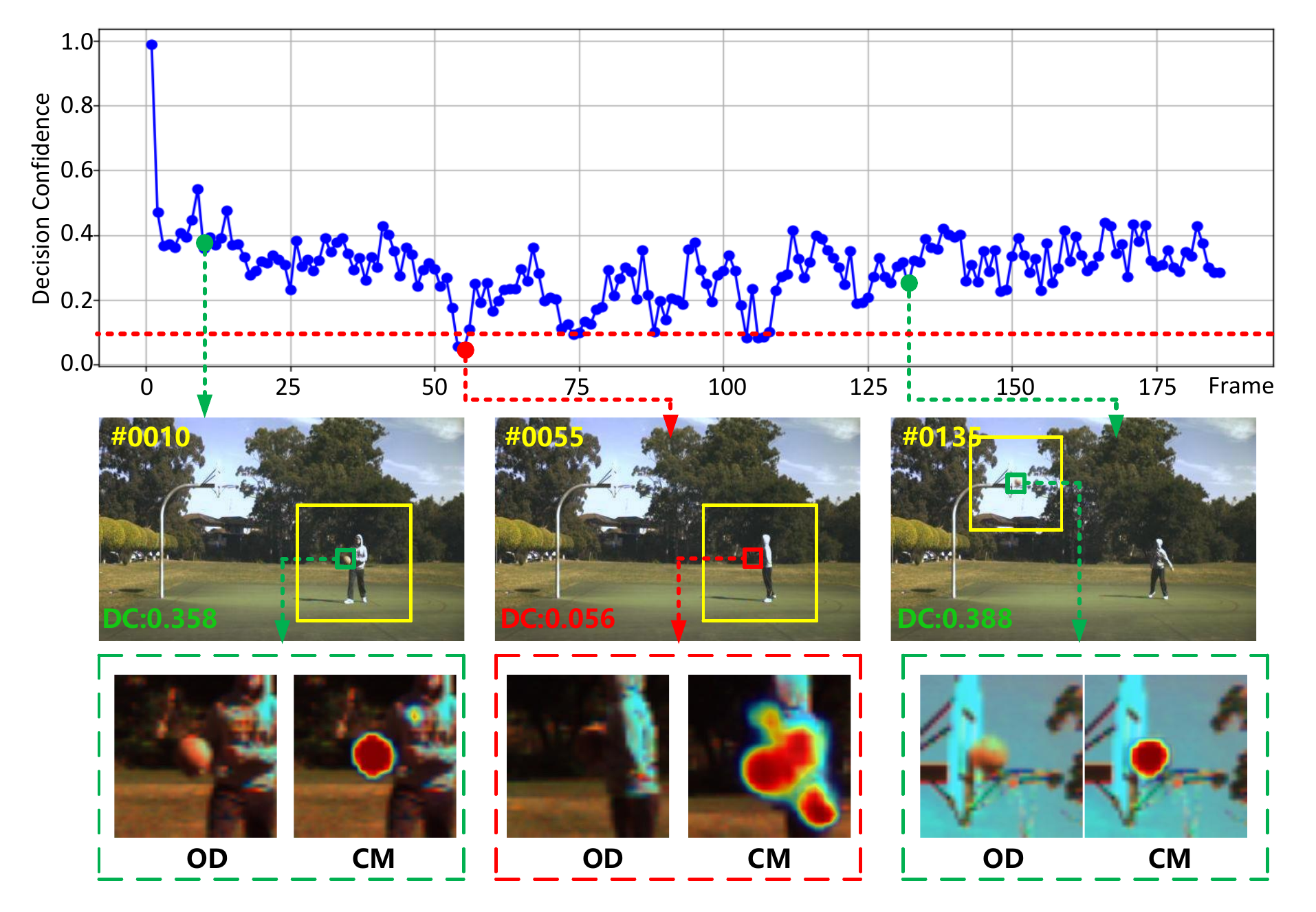}
\caption{Visualization of the decision confidence (DC) and the corresponding classification map (CM) of the \textit{basketball} video sequence on the HOTC-2020 dataset. The line graph above represents the change in DC for each frame. It can be observed that when DC is below the threshold (i.e., frame \#0055), multiple local extreme points in the CM appear, and the tracking results become unreliable. OD denotes the original images. }
\label{fig:DC}
\end{figure*}

\subsection{Distractor-aware Module}

Robust joint feature representation is crucial for tracker performance. However, the robust feature is inevitably compromised under disturbances caused by occlusion or similar background distractors, leading to the drift of tracking boxes. To address this issue, we propose the distractor-aware module (DAM), which has a three-step process: perceiving, responding, and rectifying.

\textit{Step 1. Perceiving.} Firstly, the tracker should perceive the occurrence of distractors. Inspired by \cite{li2024object}, a decision confidence metric is proposed. Specifically, we calculate the average peak-to-correlation energy of the predicted score map in the classification map, which can be formulated by
\begin{equation}
DC = \frac{{|C{M_{\max }} - C{M_{\min }}{|^2}}}{{\sum\nolimits_i^H {(C{M_i}}  - C{M_{\min }}{)^2}}},
\end{equation}
where ${C{M_{max}}}$ and ${C{M_{min}}}$ denote the maximum and minimum values of the classification map $CM$, respectively, and $H$ denotes the number of pixels of $CM$. This DC reveals the degree of fluctuation in the response map and the confidence level of the detected objects. When this value is large, it indicates that the response map is stable, and there is only one extreme point of the response value. The predicted position is close to the true one. When encountering disturbances caused by changes in conditions, the fluctuation of the response map increases, and multiple ambiguous prediction points will appear, causing a decrease in DC. Fig. \ref{fig:DC} shows the values of DC and the attention heatmap generated by the Grad-CAM++ \cite{chen2018gradnorm}. It can be observed that when there are disturbances such as occlusion and similar background interference (frame \#0055), the prediction results become unreliable. The attention heatmap shows multiple areas with high responses, resulting in ambiguous tracking results. In this way, the distractor can be perceived.


\textit{Step 2. Responding.} When a distractor occurs, causing a decrease in DC beyond the threshold $\tau$, the original tracking predicted results are misleading and may be accompanied by a mutation in positions, leading to the drift of the tracking box. Considering that, in general, the object's motion is linear and the camera imaging frequency is high (25 FPS), the object's position after the distractor depends on the motion state of the previous frame. Therefore, we utilize a Kalman filter, including the prediction and updating process, to predict the object position when a distractor occurs. The Kalman filter utilizes linear system state functions to estimate the optimal state through the observation data. It can be well adapted to HCOT tasks under distractors. Let ${s_t} = [{x_t},{y_t},{a_t},{r_t},{\dot x_t},{\dot y_t},{\dot a_t}]$ represent the state vector of objects at time $t$, where $({x_t},{y_t})$ and $({a_t},{r_t})$ denote the center coordinates of the object and the area and aspect ratio of the object, respectively, and ${\dot x_t},{\dot y_t},{\dot a_t}$ denote the variation speed of the three corresponding states. In the prediction stage, the Kalman filter focuses on solving the following state transformation equations:
\begin{equation}
\begin{array}{l}
{s_t} = F{{\hat s_{t - 1}}} + G{w_{t - 1}},\\
{E_t} = F{\hat E_{t - 1}}{F^T} + {Q_t},
\end{array}
\end{equation}
where $s_t$ denotes the prior state variable at time $t$, $\hat s_{t - 1}$ denotes the posterior state variable at time $t$, $F$ denotes the state transition matrix, $G$ denotes the state noise matrix, $w_{t - 1}$ represents the control vector of covariance matrix $Q_t$, and $E_t$ denotes the prior estimation of the error matrix. Specifically, $F$ can be formulated by
\begin{equation}
F = \left[ {\begin{array}{*{20}{c}}
1&0&0&0&1&0&0\\
0&1&0&0&0&1&0\\
0&0&1&0&0&0&1\\
0&0&0&1&0&0&0\\
0&0&0&0&1&0&0\\
0&0&0&0&0&1&0\\
0&0&0&0&0&0&1
\end{array}} \right].
\end{equation}

Given ${s_t}$, the observation variable ${z_t}$ can be obtained by
\begin{equation}
{z_t} = H{s_t} + {V_t},
\end{equation}
where ${V_t}$ denotes the observation noise variable, and $H$ denotes the observation matrix, which can be defined as
\begin{equation}
H = \left[ {\begin{array}{*{20}{c}}
1&0&0&0&0&0&0\\
0&1&0&0&0&0&0\\
0&0&1&0&0&0&0\\
0&0&0&1&0&0&0
\end{array}} \right],
\end{equation}

\begin{figure*} [!htb] 
\centering
\includegraphics[width=15.5cm,height=4.0cm]{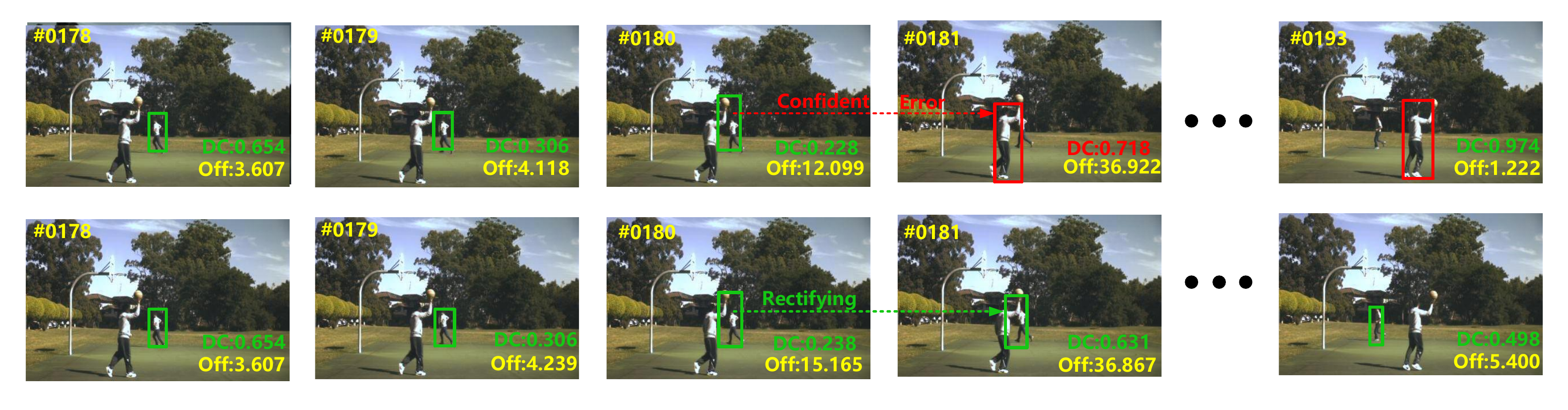}
\caption{Visualization of the “Confident Error.” In some cases of background or object confusion, the DC value is still large, and the proposed DAM cannot handle this situation. To address this issue, we propose a new rectifying strategy, which captures changes in DC before disturbances occur. Then, we can decide whether to use DAM or SPDN for box prediction based on this.}
\label{fig:DC_problem}
\end{figure*}

As for the prediction process, the posterior estimate $\hat s_{t}$ can be obtained by correcting the prior estimation $s_t$ using the observation variable ${z_t}$. The updating process can be formulated by
\begin{equation}
{k_t} = \frac{{{s_t}{H^T}}}{{H{s_t}{H^T} + {R_t}}},
\end{equation}
\begin{equation}
{{\hat s}_t} = {s_t} + {k_t}({z_t} - H{s_t}),
\end{equation}
\begin{equation}
{{\hat E}_t} = (I - {k_t}H){E_t},
\end{equation}
where $R_t$ is the covariance matrix, ${\hat s}_t$ denotes the posterior estimation of state vector, $I$ is the identity matrix, $K_t$ denotes the Kalman gain, and ${\hat E}_t$ denotes the prior estimation of the error matrix. To obtain robust tracking results further, the Kalman tracker only updates the state predicted by SPBN when the DC exceeds the threshold.

\textit{Step 3. Rectifying.} During the tracking process, SPBN combined with the Kalman filter can effectively handle short-term occlusion. However, an issue arises with the decision confidence (DC) metric proposed in \cite{li2024object}. When the environment or surrounding objects are visually similar to the tracking object, particularly when their spectral information is also similar, the DC remains high, even if the tracking box has drifted. We call this phenomenon a “Confident Error.” In such cases, the tracker trusts the tracing results from SPBN (because of the high DC), leading to incorrect tracking, as shown in Fig. \ref{fig:DC_problem}. If a perfect tracklet can be obtained, the offset of the predicted box between continuous frames remains small. Significant offsets can arise under two conditions: either a confident error has caused the tracking box to drift, or the object has been successfully re-captured after being lost due to occlusion or other distractors. Our goal is to rectify the situation with a significant offset in the first case while preserving the ability of the tracker to capture the object again in the second case. 

Different from \cite{li2024object}, we propose a rectifying strategy that combines DC and offset. Specifically, before the distractor occurs, the confidence level typically declines gradually. The tracker may incorrectly predict the position of the object because of the confident error, causing a drift of the prediction box. To address this issue, when the box offset exceeds the threshold $\psi$, we calculate the moving average of DC within the range of the past time sliding window \cite{ma2022mega} to capture the trend of confidence. A gradual decrease in DC indicates that confident errors will occur. We use the object state predicted by the Kalman filter as the tracking result to handle this situation. In this way, the problem of interference from similar objects can be addressed.

\section{Experiments}

\subsection{Experiment Settings}

\textit{1) Datasets.} In the experiment, our proposed BihoT and two HOT datasets are selected to evaluate the proposed SPDAN. The details are described as follows.

\textit{BihoT Dataset:} BihoT is captured by IMEC XIMEA MQ022HG-IM-SM5X5-NIR camera and consists of 49 HS video sequences, of which 27 are used for training and 22 are used for testing. The training part contains a total of 22,992 annotated frames, while the testing part contains 18,920 annotated frames. The dataset is annotated with 9 attributes, including Background Clutter (BC), Fast Motion (FM), In-Plane Rotation (IPR), Illumination Variation (IV), Low Resolution (LR), Occlusion (OCC), Out-of-Plane Rotation (OPR), Spectral Consistency (SC), and Spectral Variation (SV).

\textit{HOTC-2020 Dataset:} HOTC-2020 was proposed at the first Hyperspectral Object Tracking Challenge in 2020. It consists of 75 HS video sequences, of which 40 are used for training, and 35 are used for testing. The training part contains a total of 12,839 annotated frames, while the testing part contains 16,574 annotated frames.

\textit{HOTC-2023 Dataset:} HOTC-2023 was proposed at the 2023 Hyperspectral Object Tracking Challenge, which contains three sub-datasets: 15-band in the wavelength from 460 to 600 nm, 16-band in the wavelength from 600 to 850 nm, and 25-band in the wavelength from 665 to 960 nm sets. Each frame in the subset is labeled with a regression box.

\textit{2) Evaluation Metrics.} We employed two widely used evaluation metrics: the area under the curve (AUC) of the success plot and the precision rate at the threshold of 20 pixels to conduct experiments. These metrics are extensively utilized in the field to assess performance in challenging scenarios, including occlusions, and provide a robust measure of tracker accuracy and robustness. The success plots depict the percentage of frames that meet or exceed different IOU thresholds, forming a success curve with thresholds ranging from 0 to 1. The precision rate analyzes the pixel difference between the centers of the predicted and actual bounding boxes. The precision rate curve is generated by displaying the percentage of frames where the center discrepancy is less than a specified pixel threshold, which ranges up to 50 pixels. Typically, the precision plot with the threshold of 20 is used to rank the trackers, denoted as DP\_20.

\textit{3) Implementation Details.} The proposed SPDAN is implemented in PyTorch. All experiments are conducted on a system equipped with NVIDIA RTX 4090 GPU, Intel I7-13700K CPU, and 32G RAM. The backbone network SPBN and the embedding network of RGB modality are initialized with the pre-trained parameters in \cite{ye2022joint}. The AdamW optimizer with a weight decay of 0.0001 is adopted to optimize the parameters of the network. The template and search patches are set to $256 \times 256$ pixels and $128 \times 128$ pixels, respectively. The network is then trained for a total of 10 epochs. and the batch size is set to 32. Meanwhile, the thresholds for DC $\tau$ and offset $\psi$ are set to 0.1 and 20, respectively.

\begin{table}[]
\caption{ablation experimental results with the baseline model on the BihoT dataset. $\Delta$ indicates the amount of variation.}
\label{tab:ablation_bihot}
\centering
\renewcommand{\arraystretch}{1.2}
\begin{tabular}{ccccc}
\hline
Methods & AUC   & DP\_20 & $\Delta$(AUC) & $\Delta$(DP\_20) \\ \hline
B       & 0.639 & 0.767  & -             & -                \\
B+S     & 0.672 & 0.805  & +0.033        & +0.038           \\
B+D     & 0.669 & 0.795  & +0.030        & +0.028           \\
B+DA    & 0.670 & 0.799  & +0.031        & +0.032           \\
B+DA+S  & 0.677 & 0.811  & +0.038        & +0.044           \\ \hline
\end{tabular}%
\end{table}

\begin{table}[]
\caption{ablation experimental results with the baseline model on the HOTC-2020 dataset. $\Delta$ indicates the amount of variation.}
\label{tab:ablation_hotc20}
\centering
\renewcommand{\arraystretch}{1.2}
\begin{tabular}{ccccc}
\hline
Methods & AUC   & DP\_20 & $\Delta$(AUC) & $\Delta$(DP\_20) \\ \hline
B       & 0.693 & 0.924  & -             & -                \\
B+S     & 0.704 & 0.938  & +0.011        & +0.014           \\
B+D     & 0.697 & 0.929  & +0.004        & +0.005           \\
B+DA    & 0.702 & 0.935  & +0.009        & +0.011           \\
B+DA+S  & 0.725 & 0.968  & +0.032        & +0.044           \\ \hline
\end{tabular}%
\end{table}

\subsection{Ablation Experiments}

In this subsection, several ablation experiments on the proposed BihoT and HOTC-2020 datasets are conducted to evaluate the effectiveness of our methods.

\textit{1) Effectiveness of the Proposed SEN:} The spectral embedding network is proposed to refine the spectral features and form a robust prompt representation. To demonstrate the effectiveness of the SEN, we first construct a prompt-based framework following \cite{zhu2023visual}, which is denoted as “B”. Then, the SEN is utilized to incorporate the encoded spectral information into the visual features during the token embedding stage. It is denoted as “S”. We conduct the comparative experiments between the two models on the BihoT and HOTC-2020 datasets. The experimental results are listed in Tables \ref{tab:ablation_bihot} and \ref{tab:ablation_hotc20}. It can be observed that the SEN significantly improves the performance compared to the baseline model. Specifically, on the BihoT dataset, the AUC score increases by 3.3\%, and the DP\_20 improves by 3.8\%. On the HOTC-2020 dataset, the AUC score increases by 1.1\%, and the DP\_20 improves by 1.4\%. This indicates that spectral information is important for the model to capture camouflaged objects,

\textit{2) Effectiveness of the Proposed DAM:} DAM is proposed to perceive the occlusion distractors and rectify the tracking results. We add the DAM to the model to show its effectiveness. Specifically, the methods in \cite{li2024object} are denoted as “D”. The DAM is denoted as “DA”. The experimental results are listed in Tables \ref{tab:ablation_bihot} and \ref{tab:ablation_hotc20}. It can be observed that DAM further enhances the tracking performance. Specifically, it increases the AUC and DP\_20 on the BihoT dataset from 0.639 and 0.767 to 0.670 and 0.799, respectively. Compared to the baseline model, there are improvements of 3.1\% and 3.2\% in both indicators. For the HOTC-2020 dataset, DAM increases the AUC score and DP\_20 from 0.693 and 0.924 to 0.702 and 0.935, respectively. Compared with the baseline model, there are improvements of 0.9\% and 1.1\% in both indicators. These results validate the effectiveness of the proposed DAM, which can correct the tracking results misguided by the distractors through a proposed decision confidence metric and three-step processes. Furthermore, to investigate the influence of the DC threshold $\tau$, we conduct experiments on the HOTC-2020 dataset. The experimental results are listed in Table \ref{tab:apce}. We set multiple thresholds, including 0.05, 0.10, 0.15, and 0.20. It can be seen that the optimal parameter is 0.1. Under this setting, the tracker can perceive distractors accurately, effectively improving the performance of the tracker.

\begin{table}[]
\caption{ablation experimental results of different DC threshold ($\tau$). $\Delta$ indicates the amount of variation compared with the results of the first line.}
\label{tab:apce}
\centering
\renewcommand{\arraystretch}{1.2}
\begin{tabular}{ccccc}
\hline
$\tau$ & AUC   & DP\_20 & $\Delta$(AUC) & $\Delta$(DP\_20) \\ \hline
0.05   & 0.712 & 0.950  & -             & -                \\
0.10   & 0.725 & 0.968  & +0.013        & +0.018           \\
0.15   & 0.701 & 0.934  & -0.011        & -0.016           \\
0.20   & 0.695 & 0.923  & -0.017        & -0.027           \\ \hline
\end{tabular}%
\end{table}

\begin{table}[]
\caption{ablation experimental results of different embedding dimensions (Embed Dims.). $\Delta$ indicates the amount of variation compared with the results of the first line.}
\label{tab:SEN}
\centering
\renewcommand{\arraystretch}{1.2}
\begin{tabular}{ccccc}
\hline
Embed Dims.    & AUC & DP\_20 & $\Delta$(AUC) & $\Delta$(DP\_20) \\ \hline
2       & 0.651  & 0.785  & -       & -       \\ 
4        & 0.669  & 0.799  & +0.018        &  +0.014       \\ 
8     & 0.668  & 0.809  &  +0.017       &   +0.024      \\ 
16         & 0.672  & 0.805  &  +0.021       & +0.020        \\ \hline
\end{tabular}%
\end{table}

\begin{table}[]
\caption{cross-dataset validation results between BihoT and HOTC23-NIR datasets. B denotes 
 the BihoT dataset and N denotes the HOTC23-NIR dataset. “$1$ to $2$”  refers to training on dataset $1$ and testing on dataset $2$.}
\label{tab:cross_dataset}
\centering
\renewcommand{\arraystretch}{1.2}
\begin{tabular}{ccccc}
\hline
Test setting & AUC   & DP\_20 & $\Delta$(AUC) & $\Delta$(DP\_20) \\ \hline
B to B       & 0.672 & 0.805  & -                            & -                               \\
N to B       & 0.322 & 0.394  & -0.350                        & -0.411                           \\ \hline
N to N       & 0.709 & 0.901  & -                            & -                               \\
B to N       & 0.619 & 0.829  & -0.090                         & -0.072                           \\ \hline
\end{tabular}%

\end{table}

\begin{figure}
    \centering
    \includegraphics[width=7.5cm,height=5.5cm]{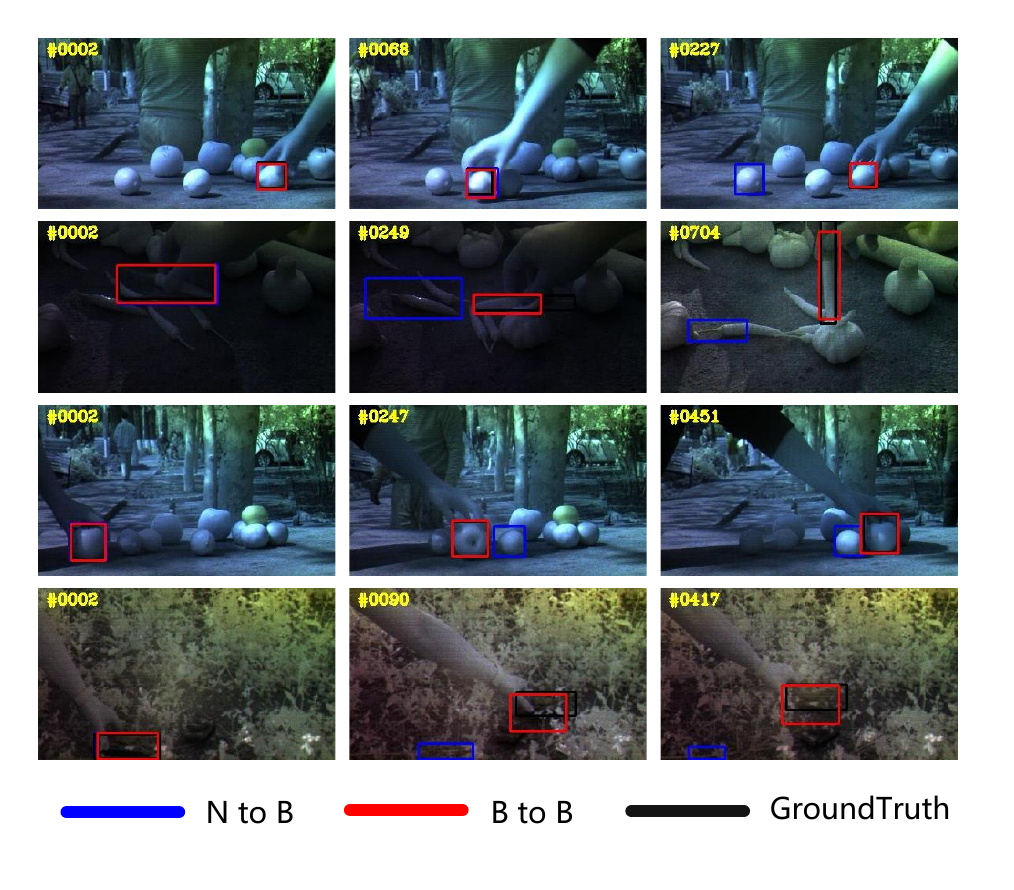}
    \caption{Visualization of the cross-dataset validation results on four video sequences (lemon2, apple3, chill2, and tank1) from the proposed BihoT dataset. “$1$ to $2$”  refers to training on dataset $1$ and testing on dataset $2$.}
    \label{fig:B2B}
\end{figure}

\textit{3) The Impact of Embedding Dimension In the SEN:} To explore the impact of embedding dimension (the number of output channels by 3-D convolutional layer) on the performance of the proposed SEN, we select multiple embedding dimension settings and conduct the tracking performance experiments on the proposed BihoT. The experimental results are listed in Table \ref{tab:SEN}. It can be seen that the four sets of embedding parameters (i.e., 2, 4, 8, and 16) are selected. Among them, the setting of embedding dimension 16 achieves the best results: the AUC score reaches 0.672, and DP\_20 reaches 0.805. Under optimal experimental settings, the SEN can capture more comprehensive discriminative information in the spectral dimension, providing richer spectral semantic information for the prompt module.

\textit{4) The Generalization Ability Validation of SPDAN.} To demonstrate the robustness of our method and the suitability of the dataset size, we perform cross-dataset validation. Specifically, our experiments involve two datasets, BihoT and HOTC23-NIR, each consisting of 25 bands. Our method entails training the SPDAN on the HOTC23-NIR training dataset and evaluating its performance on the BihoT testing dataset. Conversely, we also train the SPDAN on the BihoT training dataset and assess its performance on the HOTC23-NIR testing dataset. The experimental results are shown in Table \ref{tab:cross_dataset} and Fig. \ref{fig:B2B}. The results indicate that the model trained on the BihoT dataset performs better on the HOTC23-NIR testing dataset. This outcome serves as evidence of the generalization capability of the SPDAN model and validates the adequacy of the dataset scale for training models with broad applicability. Moreover, an interesting phenomenon is observed: the model trained on HOTC23-NIR struggles to track many camouflaged objects in the BihoT testing set. This suggests that models trained on the existing HOT dataset face challenges in handling camouflaged objects and effectively utilizing HS information to meet practical application needs, such as tracking military camouflaged objects. This underscores the necessity and feasibility of the dataset we propose.

\begin{table*}[]
\caption{overall performance comparison of HS and RGB trackers on the proposed BihoT dataset. The HS trackers deal with the HSIs, and the visual tracker runs on the false-color images. We use the red and blue fonts to highlight the top two values.}
\label{tab:bihot}
\centering
\begin{tabular}{c|ccccccc}
\hline
Trackers & OSTrack\cite{ye2022joint} & SiamBAN\cite{chen2020siamese} & MixFormer\cite{cui2024mixformer} & SEE-Net\cite{li2023learning} & SiamHYPER\cite{liu2022siamhyper} & SAM-T \cite{cheng2023segment} & SPDAN (Ours) \\ \hline
AUC      & 0.350   & 0.343   & 0.388     & 0.315   & 0.403  & \textcolor{blue}{0.478}     & \textcolor{red}{0.677} \\
DP\_20   & 0.438   & 0.445   & 0.454     & 0.422   & 0.539  & \textcolor{blue}{0.589}   & \textcolor{red}{0.811} \\ \hline
\end{tabular}%
\end{table*}

\begin{table*}[]
\centering
\renewcommand{\arraystretch}{1.2}
\setlength{\tabcolsep}{7mm}
\caption{overall performance comparison of HS and visual trackers on the HOTC-2020 dataset. The HS trackers deal with the HSIs, and the visual tracker runs on the false-color images. UFB \cite{chen2024sense} and FB denote the attempt to use full bands and feature fusion, respectively. We use the red and blue fonts to highlight the top two values.}
\label{tab:hotc20}
\begin{tabular}{ccccccc}
\hline
Tracker   & Publications & Backbone  & UFB & FB  & AUC                          & DP\_20                       \\ \hline
SiamBAN\cite{chen2020siamese}   & CVPR 2020   & ResNet-50 & No  & No  & 0.587                        & 0.863                        \\
SiamGAT\cite{guo2021graph}   & CVPR 2021   & ResNet-50 & No  & No  & 0.581                        & 0.827                        \\
SiamCAR\cite{cui2022joint}   & IJCV 2022   & ResNet-50 & No  & No  & 0.586                        & 0.846                        \\
SimTrack\cite{chen2022backbone}  & ECCV 2022   & ViT  & No  & No  & 0.600                        & 0.845                        \\
OSTrack\cite{ye2022joint}  & ECCV 2022   & ViT  & No  & No  & 0.557                        & 0.816                        \\
OSTrack*\cite{ye2022joint}  & ECCV 2022   & ViT  & No  & No  & 0.689                        & 0.914                        \\
SeqTrack\cite{chen2023seqtrack}  & CVPR 2023    & ViT  & No  & No  & 0.594                        & 0.856                        \\ 
ARTrack\cite{bai2024artrackv2}  & CVPR 2024    & ViT  & No  & No  & 0.689                        & 0.935                        \\ 
Mixformer\cite{cui2024mixformer}       & TPAMI 2024   & ViT  & No  & No  & 0.677                        & 0.934                        \\ \hline
BAE-Net\cite{li2020bae}   & ICIP 2020   & VITAL     & Yes & No  & 0.606                        & 0.879                        \\
SST-Net\cite{li2021spectral}   & WISP 2021   & VITAL     & Yes & Yes & 0.623                        & 0.917                        \\
SiamHYPER\cite{liu2022siamhyper} & TIP 2022    & SiamFC    & Yes & Yes & 0.678                        & 0.945                        \\
SEE-Net\cite{li2023learning}   & TIP 2023    & SiamFC    & Yes & No  & 0.654                        & 0.907                        \\
SiamOHOT\cite{sun2023siamohot}  & TGRS 2023   & SiamFC    & Yes & Yes & 0.634                        & 0.884                        \\
SPIRIT\cite{chen2023spirit}  & TGRS 2024   & SiamFC    & Yes & Yes & 0.679                        & 0.925                        \\
MMF-Net\cite{li2024material}   & TGRS 2024   & SiamFC    & Yes & Yes & \textcolor{blue}{0.691} & 0.932                        \\
SENSE\cite{chen2024sense}     & IF 2024     & SiamFC    & Yes & No  & 0.689                        & \textcolor{blue}{0.951} \\
SPDAN     & Ours        & ViT  & Yes & Yes & \textcolor{red}{0.725} & \textcolor{red}{0.968} \\ \hline

\end{tabular}%
\end{table*}

\subsection{Comparison with State-of-the-arts}

\begin{table*}[]
\caption{overall performance comparison of HS trackers on the HOTC-2023 dataset. We use the red and blue fonts to highlight the top two values.}
\label{tab:hotc23}
\centering
\renewcommand{\arraystretch}{1.5}
\scalebox{1.0}{
\begin{tabular}{ccccccccccc}
\hline
Band Num. & Metrics & SPIRIT   & SENSE & BAE-Net & SiamBAG & SEE-Net & SiamHYPER & SiamCAT & SPDAN (Ours) \\ \hline
\multirow{2}{*}{15}           & AUC                & 0.381 & 0.394 & 0.397   & 0.374   & 0.363   & 0.430     & \textcolor{blue}{0.480}   & \textcolor{red}{0.554}       \\ \cline{2-10} 
             & DP\_20             & 0.501 & 0.500 & 0.508   & 0.517   & 0.496   & 0.579     & \textcolor{blue}{0.665}   & \textcolor{red}{0.694}       \\ \hline
 \multirow{2}{*}{16}           & AUC                & \textcolor{blue}{0.608} & 0.607 & 0.558   & 0.554   & 0.561   & 0.590     & 0.593   & \textcolor{red}{0.669}       \\ \cline{2-10} 
                              & DP\_20             & 0.820 & 0.826 & 0.802   & 0.779   & 0.784   & 0.815     & \textcolor{blue}{0.822}   & \textcolor{red}{0.888}       \\ \hline
 \multirow{2}{*}{25}           & AUC                & 0.623 & 0.545 & 0.438   & 0.485   & 0.455   & 0.515     & \textcolor{blue}{0.665}   & \textcolor{red}{0.714}       \\ \cline{2-10} 
                              & DP\_20             & 0.838 & 0.771 & 0.779   & 0.751   & 0.753   & 0.852     & \textcolor{blue}{0.920}   & \textcolor{red}{0.923}       \\ \hline
 \end{tabular}%
 }
 \end{table*}

\textit{Performance on the BihoT Dataset.} We compare the proposed SPDAN with advanced HS and RGB trackers on the BihoT dataset, including SEE-Net \cite{li2023learning}, SiamHYPER \cite{liu2022siamhyper}, MixFormer \cite{cui2024mixformer}, SiamBAN \cite{chen2020siamese}, SAM-T \cite{cheng2023segment} and OSTrack \cite{ye2022joint}. SEE-Net reconstructs the HS bands through an autoencoder to learn the weight of each band. SiamHYPER proposes a Siamese tracking network that adopts independent networks for the extraction of features of RGB and HS modalities. MixFormer proposes a mixed attention module to extract features, and the tracking box can be obtained without a certain head network. SiamBAN proposes a box-adaptive head network to predict the accurate object positions. SAM-T amalgamates Segment Anything Model (SAM) with a proposed AOT-based tracking model to facilitate object tracking in video, which is built upon large visual models. OSTrack is a one-stage tracking framework, which extracts the features of the template and search images simultaneously through a visual Transformer-based backbone. It is worth noting that we retrained the SEE-Net and SiamHYPER according to the corresponding settings in the paper. The experimental results are listed in Table \ref{tab:bihot}. It can be observed that the proposed SPDAN achieves the best performance in both metrics. Furthermore, the BihoT dataset contains a large number of camouflage objects, which exhibit the same appearances but different spectral information. The lack of visual features poses a significant challenge to traditional RGB trackers that rely on visual features such as color and texture. This unique setting causes substantial disparities in AUC score and DP\_20 between the conventional RGB trackers and the proposed SPDAN. This discrepancy not only highlights the limitations of current trackers but also emphasizes the effectiveness of our methods. In addition, most HS trackers use band regrouping to generate three-channel false-color images and directly use RGB trackers to predict object positions without considering spectral information. They heavily rely on the performance of RGB trackers and are not conducive to improving the performance of HS tracking. However, SiamHYPER achieves commendable results. Unlike band regrouping-based methods, it extracts spectral features and fuses cross-modality features, which, to some extent, utilizes spectral information and alleviates the dependence on RGB trackers. It indicates the importance of spectral information for HOT, especially for camouflaged objects.

\textit{Performance on the HOTC-2020 Dataset.} In this sub-section, we compare the proposed SPDAN with recent HS and RGB trackers in this sub-section. HS trackers include BAE-Net \cite{li2020bae}, SST-Net \cite{li2021spectral}, SiamHYPER \cite{liu2022siamhyper}, SEE-Net \cite{li2023learning}, SiamOHOT \cite{sun2023siamohot}, SPIRIT\cite{chen2023spirit}, MMF-Net \cite{li2024material}, and SENSE \cite{chen2024sense}, which are tested on HS video sequences. The RGB trackers include SiamBAN \cite{chen2020siamese}, SiamGAT \cite{guo2021graph}, SiamCAR \cite{cui2022joint}, SimTrack \cite{chen2022backbone}, OSTrack \cite{ye2022joint}, SeqTrack \cite{chen2023seqtrack}, ARTrack \cite{bai2024artrackv2}, and MixFormer \cite{cui2024mixformer}, which are tested on false-color video sequences. The experimental results are shown in Table \ref{tab:hotc20}. OSTrack* represents conducting experiments using weights trained in 300 epochs, while OSTrack represents conducting experiments using weights trained in 100 epochs. The proposed SPDAN can achieve results of 0.725 on AUC and 0.968 on DP\_20, surpassing the above recent methods and proving the effectiveness of our method. The AUC of RGB trackers is mostly below 0.600, and the DP\_20 is below 0.900. The performance of SPDAN is superior to the methods with the same ViT backbone, proving the effectiveness of modality fusion and DAM in improving HOT performance. The performance of HS trackers is generally better than that of RGB trackers. This result proves the necessity of leveraging spectral features, as relying solely on visual features cannot handle distractors from similar objects and background clutter. Then, comparing with HS trackers, it can be observed that the use of cross-modality feature fusion performs better, indicating that fusing RGB and HS modalities can help improve performance. In addition, most existing HS trackers are based on the Siamese architecture of SiamFC \cite{bertinetto2016fully}, ignoring the advantages of the Transformer in the feature extraction. Their global relationship modeling ability is very helpful for extracting the spatial and spectral information. However, ViT-based trackers like SimTrack and SeqTrack face limitations due to their heavy reliance on visual features and a semantic discrepancy with the robust HS features, impeding their performance enhancement. Their AUC yields 0.600 and 0.594, while the DP\_20 are 0.845 and 0.856, respectively. In comparison, our SPDAN has achieved improvements of 12.5\% and 13.1\% in AUC and 12.3\% and 11.2\% in DP\_20, respectively.

\begin{figure} [!htb] 
\centering
\includegraphics[width=8.5cm,height=6cm]{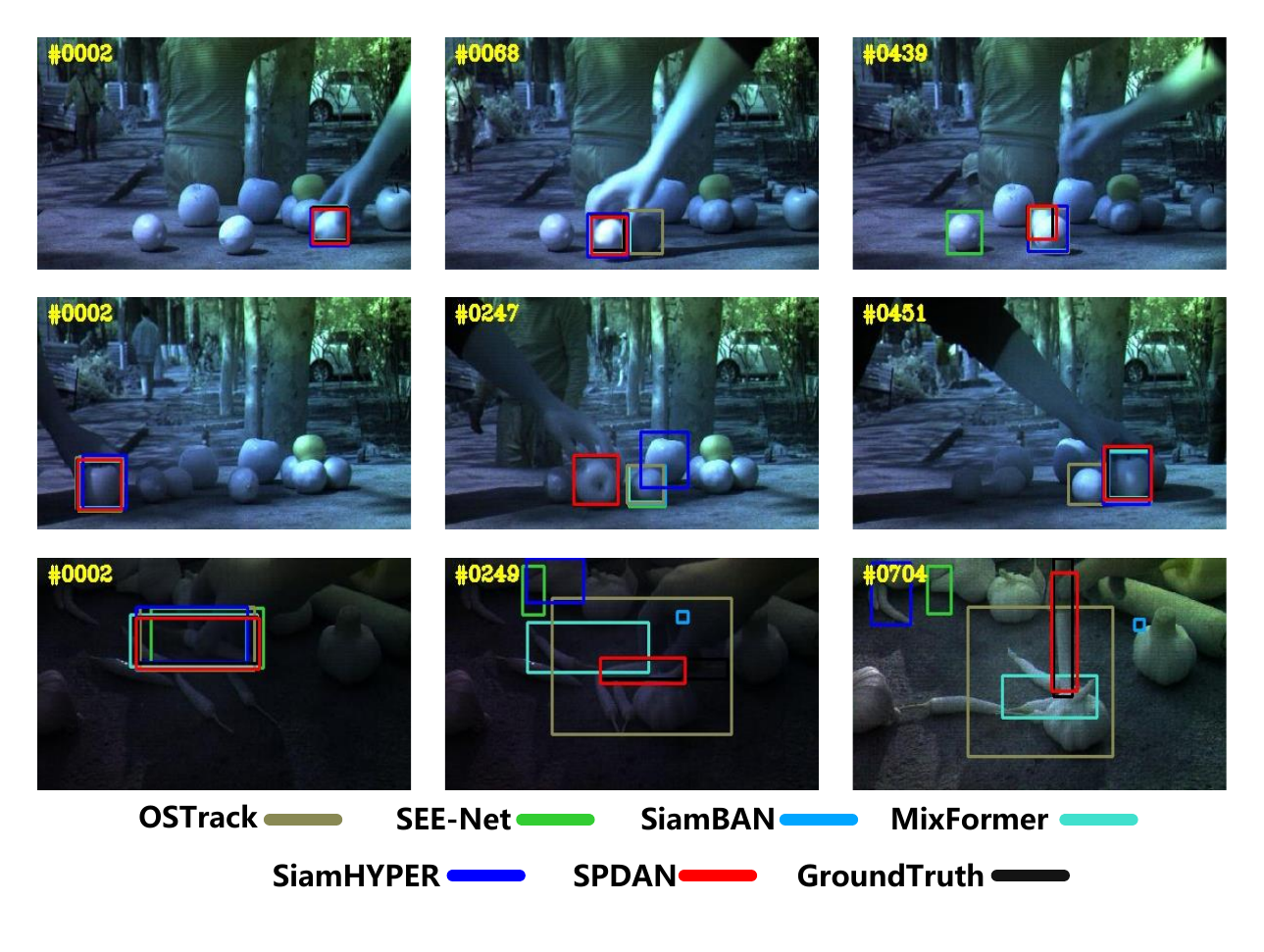}
\caption{Visualization of the tracking results on four video sequences (lemon2, apple3, and chill2) from the proposed BihoT dataset.}
\label{fig:tracking}
\end{figure}

\begin{figure} [!htb] 
\centering
\includegraphics[width=8.5cm,height=6cm]{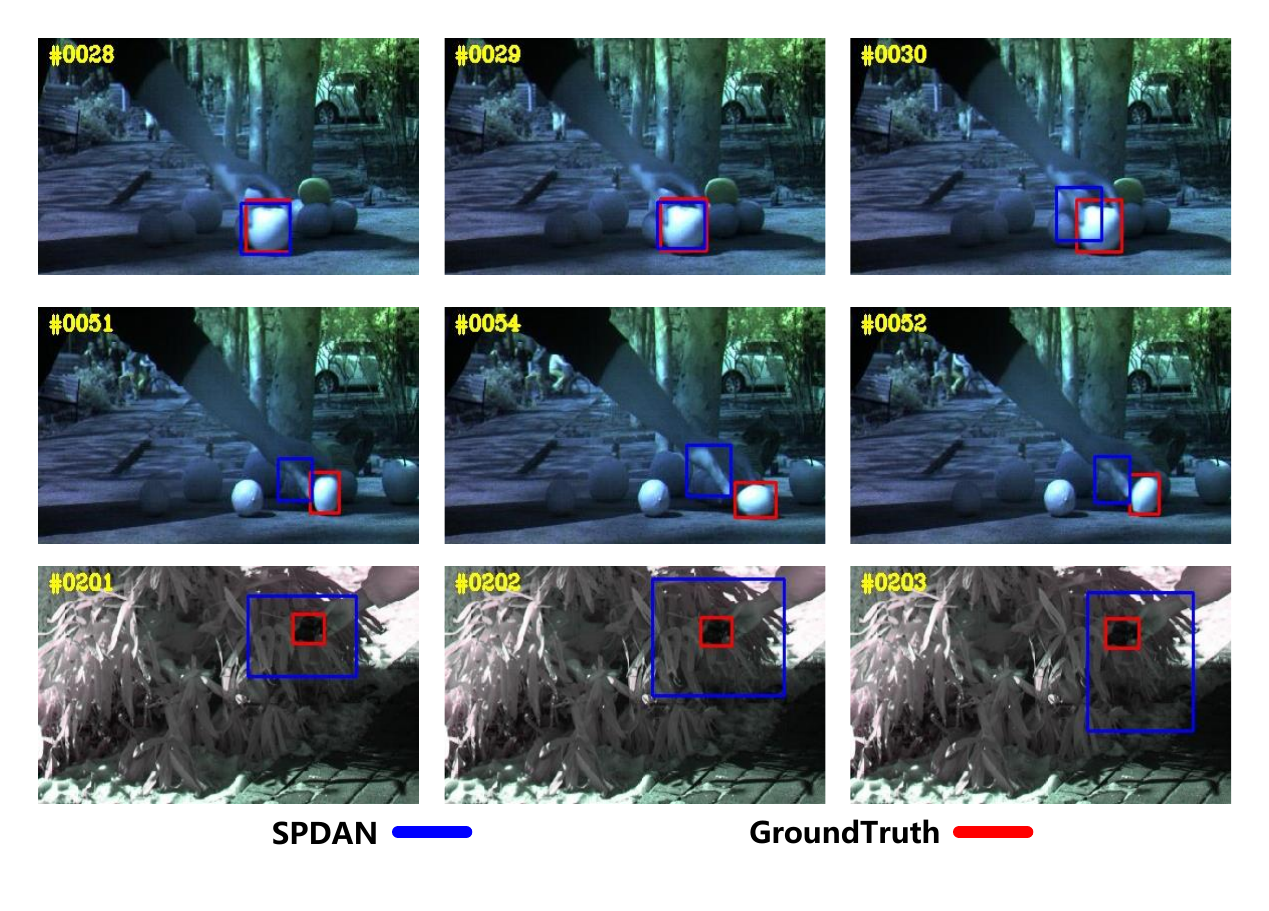}
\caption{Visualization of the failed tracking results on four video sequences (apple3, lemon3, and tank2) from the proposed BihoT dataset.}
\label{fig:defeat}
\end{figure}

\textit{Performance on HOTC-2023 Dataset.} Furthermore, to validate the effectiveness of our method, we conduct a comparative analysis of the proposed SPADN with several other trackers, including SiamCAT \cite{jiang2024channel}, SPIRIT \cite{chen2023spirit}, SENSE \cite{chen2024sense}, BAE-Net \cite{li2020bae}, SiamBAG \cite{li2023siambag}, and SiamHYPER \cite{liu2022siamhyper}, all of which are evaluated on HS images. The experimental results are listed in Table \ref{tab:hotc23}. It is evident from the findings that our SPDAN model outperforms all other trackers in tracking performance across all three bands of HS data. Notably, trackers such as BAE-Net, SiamBAG, and SEE-Net rely on band regrouping techniques using false-color images, which can result in varying image quality across bands and consequently impact tracking accuracy. In contrast, methods like SiamHYPER and our proposed SPDAN, which focus on cross-modality feature fusion, demonstrate superior results, underscoring the effectiveness of modality fusion in enhancing feature robustness.

\subsection{Visualization Analysis.}

To demonstrate the superiority of our method, we show the tracking results on the proposed BihoT dataset. We select four video sequences, including \textit{lemon2}, \textit{apple3}, \textit{chill2}, and \textit{tank1}. Due to the lack of RGB images collected in our dataset, we select the 1st, 9th, and 16th bands to form the false-color image. The experimental results are shown in Fig. \ref{fig:tracking}. The ground truth bounding box is framed by black lines. The first column is obtained from the second frame of tracking results. The results from the consecutive frames are illustrated from the second to the fourth columns. From the first and second video sequences, we can observe that most methods struggle to accurately distinguish between camouflaged and real objects, resulting in tracking failures. In contrast, our method successfully captures both visual and spectral information of the objects, demonstrating the robustness in handling camouflaged objects. In addition, for the third video sequence, SPDAN can still perform well for smaller objects, especially when dealing with object rotation. Furthermore, the fourth sequence introduces two types of challenges related to camouflage: the first is the environmental similarity to the tank model, and the second is the visual similarity between two tanks made from different materials. It can be seen that the proposed method can effectively address these two challenges. In comparison, methods like SiamHYPER and MixFormer are misled by the camouflaged tanks during object movement. Other methods are also impacted by background clutter or camouflaged elements, further underscoring the superior performance of our method. In addition, we can also see typical failure cases and challenges, as shown in Figure \ref{fig:defeat}. Due to the manual control of the object's movement, the annotation box may contain some information about the attached objects, such as hands. When disturbed by occlusion or other factors, the tracker may focus on other areas of the hand. How to extract more refined features and avoid this bias is a major challenge for future research.

\subsection{Running Cost Comparison.}

To demonstrate the superiority of our method in terms of running speed, we conduct experiments on the BihoT dataset. The experimental results using frames per second (FPS) as the evaluation indicator are reported in Table \ref{tab:fps}. All the trackers ran on a Windows machine with the configuration of an Intel(R) Core(TM) i7-13700K CPU@3.40GHz and an NVIDIA 4090 GPU. Besides, the RGB trackers ran on the three-channel false-color images, and the HS tracker ran on the HSIs. SEE-Net, SiamHYPER, and SPDAN are HS trackers, and others are RGB trackers. SPDAN achieves the best running speed among all trackers. It proves that our method has better computational efficiency compared to band regrouping or existing modality fusion methods. In addition, we also conducted experiments on computational complexity, and the experimental results are listed in Table \ref{tab:flops}. The method based on band regrouping requires multiple inferences and has higher flops. Meanwhile, the proposed SPDAN, utilizing 3D convolution, does not achieve the lowest FLOPs or model size but represents a balanced trade-off for performance improvement.

\begin{table}[]
\caption{Running time comparison in terms of frames per second (FPS) on the BihoT datasets. The RGB trackers ran on the three-channel false-color images, and the HS tracker ran on the HSIs.}
\label{tab:fps}
\centering
\resizebox{\columnwidth}{!}{%
\begin{tabular}{cccccc}
\hline
Tracker        & SiamBAN\cite{chen2020siamese} & MixFormer\cite{cui2024mixformer} & SEE-Net\cite{li2023learning} & SiamHYPER\cite{liu2022siamhyper} & SPDAN \\ \hline
False-color     & 15.77  & 18.54     & n/a     & n/a       & n/a   \\
Hyperspectral    & n/a     & n/a       & 9.27    & 16.59     & 21.02 \\ \hline
\end{tabular}%
}
\end{table}

\begin{table}[]
\caption{Computation complexity comparison results including FLOPs and model size.}
\label{tab:flops}
\resizebox{\columnwidth}{!}{%
\begin{tabular}{cccccc}
\hline
Methods  & SEE-Net\cite{li2023learning}  & SiamBAG\cite{li2023siambag}  & TransT\cite{chen2021transformer}  & ARTrack\cite{bai2024artrackv2} & SPDAN \\ \hline
FLOPs    & 297.992G & 109.277G & 19.194G & 37.150G  & 37.781G  \\
Size     & 53.938M  & 90.442M  & 18.542M & 173.105M  & 142.702M \\ \hline
\end{tabular}%
}
\end{table}

\begin{table*}[]
\caption{Attribute-based tracking results in terms of AUC and DP\_20 on the Bihot dataset. The best result is highlighted in bold.}
\label{tab:bihot_att}
\centering
\renewcommand{\arraystretch}{1.2}
\scalebox{0.85}{
\begin{tabular}{c|ccccccccc|ccccccccc}
\hline
\multirow{2}{*}{Trackers} & \multicolumn{9}{c|}{AUC}                                                                                                                               & \multicolumn{9}{c}{DP\_20}                                                                                                                             \\ \cline{2-19} 
                          & BC             & FM             & IPR            & IV             & LR             & OCC            & OPR            & SC             & SV             & BC             & FM             & IPR            & IV             & LR             & OCC            & OPR            & SC             & SV             \\ \hline
OSTrack                  & 0.198          & 0.271          & 0.283          & 0.341          & 0.150          & 0.363          & 0.316          & 0.222          & 0.385          & 0.271          & 0.375          & 0.352          & 0.440          & 0.179          & 0.442          & 0.387          & 0.296          & 0.474          \\
SiamBAN                   & 0.154          & 0.220          & 0.303          & 0.261          & 0.220          & 0.369          & 0.373          & 0.238          & 0.394          & 0.215          & 0.301          & 0.383          & 0.329          & 0.307          & 0.474          & 0.487          & 0.305          & 0.514          \\
MixFormer                 & 0.225          & 0.259          & 0.318          & 0.286          & 0.300          & 0.420          & 0.395          & 0.253          & 0.451          & 0.262          & 0.337          & 0.377          & 0.342          & 0.330          & 0.490          & 0.465          & 0.292          & 0.533          \\
SEE-Net                & 0.191          & 0.209          & 0.236          & 0.283          & 0.249          & 0.324          & 0.317          & 0.252          & 0.353          & 0.273          & 0.310          & 0.310          & 0.385          & 0.360          & 0.429          & 0.487          & 0.346          & 0.471          \\
SiamHYPER                & 0.294          & 0.131          & 0.339          & 0.344          & 0.465          & 0.420          & 0.411          & 0.261          & 0.457          & 0.410          & 0.204          & 0.444          & 0.445          & 0.628          & 0.553          & 0.563          & 0.348          & 0.617          \\
SPDAN (Ours)                     & \textbf{0.497} & \textbf{0.745} & \textbf{0.720} & \textbf{0.701} & \textbf{0.577} & \textbf{0.738} & \textbf{0.693} & \textbf{0.627} & \textbf{0.702} & \textbf{0.623} & \textbf{0.898} & \textbf{0.855} & \textbf{0.854} & \textbf{0.724} & \textbf{0.877} & \textbf{0.826} & \textbf{0.789} & \textbf{0.830} \\ \hline
\end{tabular}%
}
\end{table*}

\subsection{Attribute-Based Evaluation.}

To demonstrate the ability to handle different challenge scenarios of our proposed SPDAN, we conduct attribute-based experiments with the selected SOTA trackers on the BihoT datasets. The experimental results are listed in Table \ref{tab:bihot_att}. It is worth noting that to make more appropriate attribute comparisons, we redefine the challenge scenarios based on the characteristics of the BihoT, as detailed in Table \ref{tab:2}. Unlike the settings in HOTC-2020, we remove attributes (i.e., DEF, MB, and OV) and introduce scenarios such as SC (Spectral Consistency) and SV (Spectral Variability) to more accurately assess the utilization of spectral information. Consequently, the number of attributes evaluated is reduced from 11 to 9. It can be seen that our SPDAN exhibits superior performance in different scenarios, which further proves the effectiveness of our proposed method. Specifically, in the two spectral-related scenarios SC and SV, SPDAN achieves an AUC of 0.627 and 0.702, and DP\_20 of 0.789 and 0.830. The proposed method can accurately predict the position of the object in the presence of visual similarity, spectral differences, and spectral similarity interference, demonstrating that our spectral embedding network and prompt-based learning method can accurately extract spectral features for accurate tracking.

\section{Conclusion}

In this paper, we proposed a hyperspectral camouflaged object tracking task and constructed a large-scale HCOT dataset, BihoT, which contains 41,912 hyperspectral images where objects have similar appearances, diverse spectrums, and frequent occlusion. Then, we proposed a spectral prompt-based distractor-aware network (SPDAN) as a simple yet effective baseline, which combines discriminative information from the HS and RGB modalities through prompt-based learning. A spectral embedding network was designed to transform the spatial and spectral information into tokenized features. To make the tracking results more robust, we introduced a distractor-awareness module to perceive the distractor and rectify the original results. The quantitive and qualitative experimental results demonstrated the effectiveness of our proposed method. In future work, more attention will be given to leveraging temporal information in video sequences to enhance HOT performance. By integrating time-series data, trackers may better capture object motion patterns and improve robustness in challenging scenarios such as occlusions or rapid appearance changes.

\ifCLASSOPTIONcaptionsoff
  \newpage
\fi

\bibliographystyle{IEEEtran}
\bibliography{IEEEabrv,bare_jrnl}

\end{document}